%% file: paper.tex
\documentclass[12pt,onecolumn]{IEEEtran}

\usepackage[final]{graphicx}
\usepackage{amsmath}
\usepackage{amssymb}
\usepackage{algorithmic}
\usepackage{url}
\usepackage{caption}
\usepackage{subcaption}
\usepackage{tikz}
\usepackage{hyperref}
\usepackage{xcolor}
\usepackage{empheq}
\usepackage{fancybox}

\hypersetup{
    colorlinks,
    citecolor=green,
    filecolor=black,
    linkcolor=blue,
    urlcolor=blue
}

\newcommand{\ve}{\mathbf{e}}

\newcommand{\vu}{\mathbf{u}}
\newcommand{\vv}[1]{\mathbf{#1}}

\newcommand{\vx}{\mathbf{x}}

\newcommand{\mA}{\mathbf{A}}

\newcommand{\mD}{\mathbf{D}}

\newcommand{\mG}{\mathbf{G}}

\newcommand{\mI}{\mathbf{I}}

\newcommand{\mK}{\mathbf{K}}
\newcommand{\mL}{\mathbf{L}}

\newcommand{\mR}{\mathbf{R}}

\newcommand{\argmax}[1]{\operatorname*{arg\,max}_{#1}}
\newcommand{\argmin}[1]{\operatorname*{arg\,min}_{#1}}

\newcommand{\tr}{\operatorname{tr}}

\newcommand{\ltriplebar}{\lvert\kern-0.9pt\lvert\kern-0.9pt\lvert}
\newcommand{\rtriplebar}{\rvert\kern-0.9pt\rvert\kern-0.9pt\rvert}

\newcommand{\bb}{\mathbf{b}}

\newcommand{\bu}{\mathbf{u}}

\newcommand{\bA}{\mathbf{A}}

\newcommand{\bD}{\mathbf{D}}

\newcommand{\bH}{\mathbf{H}}
\newcommand{\bI}{\mathbf{I}}

\newcommand{\bK}{\mathbf{K}}
\newcommand{\bL}{\mathbf{L}}

\newcommand{\bR}{\mathbf{R}}

\newcommand{\bW}{\mathbf{W}}

\definecolor{PlotColorA}{HTML}{1f77b4}
\definecolor{PlotColorB}{HTML}{ff7f0e}
\definecolor{PlotColorC}{HTML}{2ca02c}
\definecolor{PlotColorD}{HTML}{d62728}
\definecolor{PlotColorE}{HTML}{9467bd}
\definecolor{PlotColorF}{HTML}{8c564b}
\definecolor{PlotColorG}{HTML}{e377c2}

\begin{document}

\title{Local Kernels that Approximate Bayesian Regularization and Proximal Operators\\ }
\author{Frank Ong, Peyman Milanfar, Pascal Getreuer \\ Google Research}
\date{}

\maketitle

\begin{abstract}
In this work, we broadly connect kernel-based filtering (e.g. approaches such as
the bilateral filters and non-local means, but also many more) with general
variational formulations of Bayesian regularized least squares, and the related
concept of proximal operators. The latter set of variational/Bayesian/proximal
formulations often result in optimization problems that do not have closed-form
solutions, and therefore typically require {\em global} iterative solutions. Our
main contribution here is to establish how one can approximate the solution of
the resulting global optimization problems with use of locally adaptive filters
with specific kernels. Our results are valid for small regularization strength
but the approach is powerful enough to be useful for a wide range of
applications because we expose how to derive a ``kernelized'' solution to these
problems that approximates the global solution in one-shot, using only local
operations. As another side benefit in the reverse direction, given a local
data-adaptive filter constructed with a particular choice of kernel, we enable
the interpretation of such filters in the variational/Bayesian/proximal
framework.

% In particular using a convex regularization function positive-definite kernel,
%For each kernel $k(t)$, we show that we can find a regularization function,
%expressed in terms of $\iint k(t') \mathrm{d} t'$, such that the regularized
%least squares solution approximates the filtered output. Our proposed
%regularization function is convex and smooth, thus can be minimized efficiently
%using standard smooth convex optimization algorithms. The proposed
%approximation allows us to obtain explicit signal priors for kernel methods by
%interpreting them as maximum a posteriori (MAP) estimators.
\end{abstract}


%-------------------------------------------------------------------------------
\section{Introduction}

Kernel-based non-linear filtering, such as bilateral
filter~\cite{smith1997susan, tomasi1998bilateral} and non-local
means~\cite{buades2005review}, is one of the most effective and popular methods
used in modern image processing~\cite{milanfar2013tour}. Besides denoising/smoothing,
kernel methods have shown successes in many other tasks, including image
enhancement~\cite{talebi2014nonlocal},
super-resolution~\cite{protter2009generalizing}, and
demosaicing~\cite{buades09}. Recent techniques~\cite{isidoro2016pull,
talebi2016fast} have also been proposed to speed up kernel methods, making them
feasible to run on mobile devices.

On the other hand, Bayesian/variational/proximal methods have the advantage of well-defined objective
functions, and allow the direct incorporation of penalty terms to control the
properties of the resulting solutions. These estimators use explicit signal
priors within the frameworks of maximum a posteriori (MAP) or minimum
mean-squared error (MMSE). Moreover, the statistically based methods
can easily be adapted to more general measurement models by changing the data
fidelity term. However, a major drawback is that the solution often requires
iterative global optimization methods, and is rarely used in practical situations where
speed is not only desired, but absolutely required.

his is not the first time the relationship between local filters and global
regularization has been considered \cite{elad2002origin,takeda07}. Among the
many such works, three are most directly related to ours: Durand et
al.~\cite{durand2002fast} showed the relation between bilateral filters and
robust statistics. Namely, that the bilateral is related to the first gradient
step of minimizing a cost function with a specific robust regularization
function. Durand's work built on the elegant work of Black et al.
in~\cite{black1998robust} where they showed connections between three techniques
for image filtering: anisotropic diffusion, robust statistics, and
regularization. These connections made it possible to analyze, design,
implement, and interpret anisotropic diffusion using the tools of robust
statistics. Finally, in the relatively recent work of Louchet et
al.~\cite{louchet2011total}, they proved (in the specific context of Total
Variation (TV)) that the local interactions induced by TV do not propagate much
at long distances in practice, so that the TV filtering model can be adequately
interpreted and implemented as a local filter. In fact, they built a purely
local filter by considering the TV model in a given neighborhood of each pixel,
and illustrated that for small regularization parameters, it behaved very
closely to the ideal global TV.

What is the contribution then of the present work in light of these illuminating previous works?
\begin{itemize}
\item With respect to~\cite{louchet2011total}, we aim to illustrate that the connections between local filtering and global optimization via a variational formulation go far beyond Total Variation, encompassing a much wider class of regularization functions.
\item The significant speedups enabled by the approach proposed in Durand's work~\cite{durand2002fast} subsequently enabled a fantastically wide range of use cases for the bilateral filter, making it a standard tool in both vision and graphics. Our work aims to show that by localizing a much wider range of complex Bayesian/variational filters, similar speedups may be realized for much more sophisticated filters beyond the bilateral.
\item The work of Black et al.~\cite{black1998robust} generalized the notion of anisotropic diffusion using tools from robust statistics, and made a connection to image filtering. By contrast our work takes the somewhat wider and more practical view that a more or less arbitrary regularization function (robust of otherwise) can be converted or approximated with a localized (or ``kernelized'') image filter. In the process, the notion of anisotropic diffusion is naturally captured as iterations of the derived local filters with small steps, as explained in~\cite{milanfar2013tour}.
\item Finally, our approach makes the novel case that very general variational restoration formulations can be directly associated to diffusion by local filters through the (graph) Laplacian operator. This is significant and timely with respect to the wide use of proximal operators in learning, optimization theory, and their particular use in solving  inverse problems \cite{plugandplay,meinhardt17learning,Combettes2011,condat2014,romano2016little}.  Approaches such as \cite{plugandplay} have become popular for suggesting an ad hoc procedure for replacing the proximal operators with simpler (often patch-based) denoisers. The current work is a more principled approach toward that replacement, and makes the calculation of the appropriate kernel filters more direct and transparent.

%In particular, this connection is possible at two levels which involved either the gradient of the loss (first order) or the Hessian of the loss (second order). We illustrate that the first order approach, consistently with earlier literature, connects kernels to loss functions through the {\em influence functions} of the loss. This approach does not guarantee the correspondence of positive definite kernels to convex loss functions. Whereas in the second order approach, we can ensure that the kernels do in fact correspond to positive definite Hessian of the loss and therefore guarantee its convexity.
\end{itemize}

%-------------------------------------------------------------------------------
\section{Background}

Removing noise from a signal may seem like a solved problem
\cite{chatterjee11,talebi2013saif,milanfar2013symmetrizing,chatterjee2010denoising}, but it is still an ongoing area of work. More importantly, denoising (i.e. smoothing) is a fundamental operation in all of image processing,  statistics, and machine learning. It is not only this wide use, but the fundamental nature of the denoising problem and its relationship to proximal operators that motivates our presentation here.  This said, while denoising is the appropriate platform on which to illustrate the main concepts proposed in this paper, it is important to note that the results of this paper are not meant for only denoising applications per se. As we have illustrated elsewhere~\cite{romano2016little,talebi2016fast}, insight on how to implement, approximate, and generalize denoisers is very widely useful not only as a way to describe proximal operators, but also as a key building block to nearly all facets of image processing more generally.

Now, let's specifically consider the measurement $\vx$ of a clean signal
$\vu$, corrupted by zero-mean Gaussian white noise of variance $\sigma^2$,
as\footnote{More generally, the inverse problem may involve an additional
(e.g. linear) operator acting on the clean image, so that in the measurement model
(\ref{measurement_model}) $\vu$ may be replaced by $\bH\vu$, where $\bH$ can
encode the effect of some linear degradation model such as blur. To maintain
focus, we take $\bH=\bI$, the identity operator.}
\begin{equation}\label{measurement_model}
\vv{x} = \vv{u} + \vv{e}.
\end{equation}
We note that this model is an adequate description for images scanned lexicographically into vectors.  The aim of
any denoiser is to recover an estimate of the clean underlying signal as a
function applied to the noisy measurement,
\begin{equation}
\vv{\Hat{x}} = f(\vv{x}).
\end{equation}
The denoiser is a map $f: \mathbb{R}^N \rightarrow \mathbb{R}^N$. Denoisers
enjoy many interesting functional properties, which are elaborated and
extensively used in \cite{romano2016little}.

\subsection{MAP Denoising and Proximal Operators}
Bayesian denoising invokes the use of a prior $P(\vu)$ on the class of ``clean''
images $\vu$. This prior influences the estimate of the underlying signal away
from the direct measurement $\vx$. In particular, (as the name indicates) the
maximum a posteriori (MAP) estimate is the value of $\vu$ at which the posterior
density $P(\vu|\vx)$ is maximized,
\begin{equation}
\vv{\Hat{x}}_\mathit{map} = \argmax{\vu}P(\vu|\vx).
\end{equation}
When the noise is Gaussian and white (as we have assumed), the optimization
boils down to a regularized least-squares problem
\begin{equation}
\vv{\Hat{x}}_\mathit{map}
= \argmin{\vu}\frac{1}{2\sigma^2}||\vu-\vx||^2  + \phi(\vu)
\end{equation}
where $\phi(\vv{u}) = -\log P(\vv{u})$ is the negative log-prior on the space of
clean signals. This expression for the MAP estimate coincides precisely
with what is known in the optimization literature as the {\em proximal} operator
~\cite{proximal2015} for $\phi$. Proximal operators have many interesting
properties of their own. In this exposition, however, we simply mention them
here and leave the further exploration of these connections for another
occasion.

Let's develop a more explicit form for this solution by computing
the gradient of the right-hand-side and solving,
\begin{align}
\nabla_{\mathbf{u}} \left[\frac{1}{2\sigma^2}||\vu-\vx||^2
+ \phi(\vu) \right] &= 0 \\
\frac{1}{\sigma^2}(\vu-\vx)  + \nabla\phi(\vu) &= 0 \\
\vu + \sigma^2\nabla\phi(\vu) &= \vx \label{e:map_euler_lagrange}\\
(\mI + \sigma^2\nabla\phi)(\vu) &= \vx
\end{align}
In the last line above, if we consider $\mI + \sigma^2\nabla\phi$ as an
operator acting on its argument, we can (at least symbolically, if not
explicitly) invert the operator to obtain
\begin{equation}
\vv{\Hat{x}}_\mathit{map} = (\mI + \sigma^2\nabla\phi)^{-1} (\vx).
\end{equation}
This expression is known as the \emph{resolvent}~\cite{proximal2015} of the
operator $\nabla\phi$. When $\sigma^2$ is small, we can approximate the
resolvent
\begin{equation}
(\mI  + \sigma^2\nabla\phi)^{-1} \approx \mI - \sigma^2\nabla\phi
\end{equation}
The left hand side, the MAP solution, can be interpreted as an implicit Euler
step of size $\sigma^2$ on the differential equation $\partial_t \vv{u} =
-\nabla\phi(\vv{u})$. The approximation on the right hand side is an explicit
Euler step, and we can clearly see the ``shrinkage'' effect of the denoiser,
where the second term shrinks the measured noisy signal toward zero by an amount
proportional to the gradient of the loss function $\phi(\vv{x})$. The MAP
estimate is therefore approximated by
\begin{equation} \label{eq:MAP_Shrinkage}
\vv{\Hat{x}}_\mathit{approx} = \vx - \sigma^2\nabla\phi(\vx).
\end{equation}
We can estimate the error in this approximation by letting $\vu = \widehat{\vx}_{map}$ and observing from (\ref{e:map_euler_lagrange}) that
\begin{eqnarray}
\vv{\Hat{x}}_\mathit{map} - \vv{x} & = -\sigma^2 \nabla
\phi(\vv{\Hat{x}}_\mathit{map}),
\end{eqnarray}
while
\begin{eqnarray}
\vx  =  \widehat{\vx}_{approx} + \sigma^2 \nabla\phi(\vx)
\end{eqnarray}
Therefore:
\begin{equation}
\begin{aligned}
\vv{\Hat{x}}_\mathit{map} - \vv{\Hat{x}}_\mathit{approx}
&= \vv{\Hat{x}}_\mathit{map} - \vv{x}
+ \sigma^2 \nabla \phi(\vv{x}) \\
&= -\sigma^2 \nabla \phi(\vv{\Hat{x}}_\mathit{map})
+ \sigma^2 \nabla \phi(\vv{x}) \\
&= \sigma^2 \bigl(
\nabla \phi(\vv{x})
- \nabla \phi(\vv{\Hat{x}}_\mathit{map}) \bigr).
\end{aligned}
\end{equation}
Thus the approximation relies on one or both of $\sigma^2$ and $\|\nabla
\phi(\vv{x}) - \nabla \phi(\vv{\Hat{x}}_\mathit{map})\|$ being small. If for
instance $\nabla\phi$ is Lipschitz with constant $M$, then
\begin{equation}
\|\vv{\Hat{x}}_\mathit{map} - \vv{\Hat{x}}_\mathit{approx} \|
\le \sigma^2 M \|\vv{\Hat{x}}_\mathit{map} - \vv{x}\| ,
\end{equation}
so the error is small if the denoising residual $\vv{\Hat{x}}_\mathit{map}
- \vv{x}$ is small.  For $\nabla \phi$ a linear operator, $M$ is the spectral
norm (largest singular value) of $\nabla \phi$.  Alternatively if $\phi$ is
twice differentiable, then the error can be bounded by the mean value theorem in
terms of the max spectral norm of the Hessian $\mathcal{H}\phi$ along the line
segment connecting $\vv{x}$ and $\vv{\Hat{x}}_\mathit{map}$,
\begin{equation}
\|\vv{\Hat{x}}_\mathit{map} - \vv{\Hat{x}}_\mathit{approx} \|
\le \sigma^2 m \|\vv{\Hat{x}}_\mathit{map} - \vv{x}\|
\end{equation}
where
\begin{equation}
m = \sup_{0 < c < 1}
\bigl\|\mathcal{H}\phi\bigl((1 - c) \vv{\Hat{x}}_\mathit{map} + c \vv{x}
\bigr)\bigr\|.
\end{equation}

\begin{figure}
\centering
\includegraphics[width=4in]{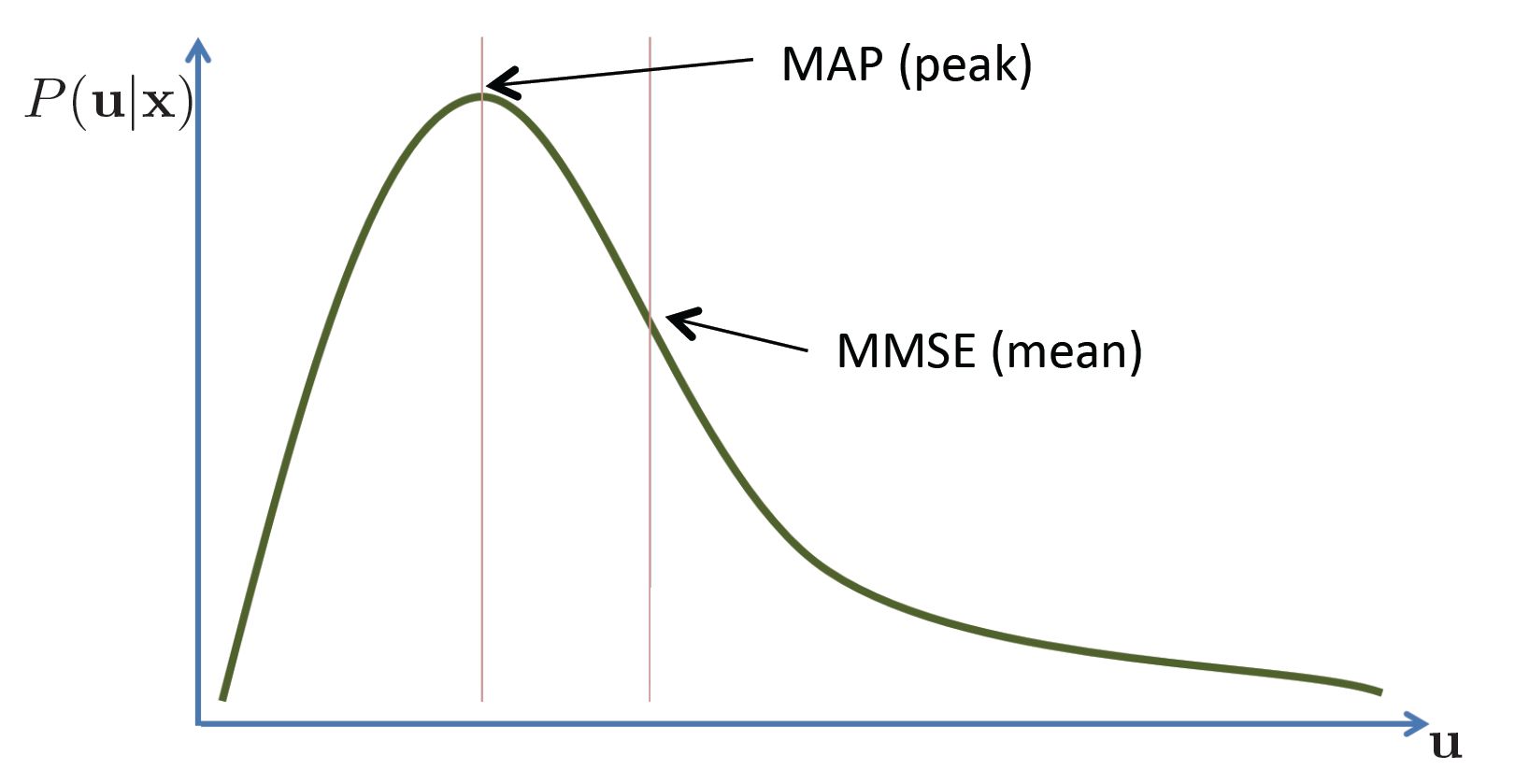}
\caption{Bayesian Estimators: MAP vs.\ MMSE}
\label{fig:BayesEstimators}
\end{figure}

We will next consider the MMSE framework, which instead of the maximum of the
posterior, considers its mean. This contrast between the MAP and MMSE is
highlighted in Figure \ref{fig:BayesEstimators}. It is interesting to note from
this cartoon that the two estimates will tend to coincide when the posterior is
symmetric and unimodal, or when the noise variance $\sigma^2$ is small. The
latter condition is interesting to the extent that it allows us to formulate our
results in a unified way across both MAP and MMSE estimators. Before we do this,
let's examine the form and function of the MMSE denoisers in a bit of detail.

\subsection{MMSE Denoising}
In contrast to MAP denoising, MMSE denoising relies on the {\em mean} of the
posterior density,
\begin{equation}
\vv{\Hat{x}}_\mathit{mmse} = \mathbb{E}(\vu|\vx) = \int \vu P(\vu|\vx) \, d\vu.
\end{equation}
The MMSE expectation integral is often impractical to evaluate directly, especially in high dimensions.
But there is a clever way around this difficulty, due to a classical result known as Tweedie's formula \cite{robbins56,stein81,efron11},
\begin{align} \label{eq:MMSE}
\vv{\Hat{x}}_\mathit{mmse} &= \mathbb{E}(\mathbf{u}|\mathbf{x}) \\
&= \vx + \sigma^2 \frac{\nabla P(\vx)}{P(\vx)}  \\
&= \vx + \sigma^2 \nabla \log P(\vx),
\end{align}
where $P(\vx)$ is the marginal density of the measurement $\vx$, which is
computed from the prior $P(\vv{u})$ as $P(\vv{x}) = \int P(\vv{x}|\vv{u})
P(\vv{u}) \,d\vv{u}$.
% The above formula is remarkable in that the MMSE estimate apparently does not
% directly depend on the prior $P(\vu)$.  Instead, it only depends on the marginal
% density of the measurement $P(\vx)$!
This marginal density is a ``blurred'' version of the prior because when $\ve$ is
Gaussian white noise of variance $\sigma^2$,
\begin{equation}
\vv{x} = \vv{u} + \vv{e} \quad \implies \quad P(\vx) = \int
\frac{\exp(-\|\mathbf{x-u}\|^2/2\sigma^2)}
{\sqrt{2\pi \sigma^2}}
 \, P(\mathbf{u}) \, d\mathbf{u}.
\end{equation}
That is, the marginal $P(\vx)$ is the convolution of the prior $P(\vv{u}) \propto
\exp\bigl(-\phi(\vv{u})\bigr)$ with a Gaussian blur centered at zero, with
variance $\sigma^2$. The MMSE expression can be further written as
\begin{equation} \label{eq:MMSE_Shrinkage}
\vv{\Hat{x}}_\mathit{mmse} = \vx - \sigma^2 \nabla \phi_\sigma(\vv{x})
\end{equation}
where $\phi_\sigma(\vv{x}) = -\log(P(\vv{x}))$. That is, $\phi_\sigma(\vv{x})$
is the negative log of the blurred prior. We can directly compare the above
expression for MMSE to the expression (\ref{eq:MAP_Shrinkage}) for the MAP
estimator and note that they are similar. Indeed, the MMSE estimate is also a
shrinkage of $\vx$ toward zero, and a gradient descent step on (a blurred
version of) the penalty $\phi(\vv{x})$. This observation underscores the
importance of the \emph{shrinkage} idea in all forms of Bayesian/variational/proximal smoothing. Going forward, we will consider the following form as the fundamental equation for all such formulations of smoothing. Namely
\begin{equation} \label{eq:bayes_shrinkage}
\vv{\Hat{x}} = \vx - \sigma^2 \nabla\phi(\vx)
\end{equation}
where $\phi(\vx)$ may henceforth be interpreted as a log-prior, a smoothed
log-prior, or more generally an arbitrary but (almost everywhere differentiable)
regularization function that may be used in a proximal setting.

\subsection{Kernel Filters}
A fundamentally different approach to smoothing is to compute data-dependent
weighted averages of the input signal values to yield each element of the
output. As described below, the affinity (or similarity) of pixels is measured using a ``kernel'' (a symmetric function), which when normalized, yields the weights for the above averaging. This framework (sometimes referred to as kernel regression
\cite{milanfar2013tour}) does not rely explicitly on a prior as does the
Bayesian framework, though as shown in \cite{milanfar2013tour} the kernel
approach can be interpreted as an \emph{empirical} Bayes procedure.  The
resulting kernel filters we shall concern ourselves with are \emph{pseudo}-linear.
Specifically, estimated pixels are computed from the measured pixels by
\begin{equation}
\Hat{x}_i=\sum_j W_{ij} \:x_j
\end{equation}
with the important caveat that the weights $W_{ij}$ can depend on $\vx$.
To be more compact with our notation, we can gather
all weights corresponding to the $i$th estimated pixel as a row vector
\begin{equation}
\vv{w}_{i}^T  = \left[ W_{i1},\:W_{i2},\:\cdots,\:W_{in}\right].
\end{equation}
And collecting all these weights into a matrix we have
 \begin{equation}
\bW = \left[ \begin{array}{c} \mathbf{w}_{1}^T \\
\vdots \\  \mathbf{w}_{N}^T \end{array} \right],
\end{equation}
from which the denoiser can be conveniently written in {\em pseudo-linear} form as
\begin{equation} \label{eq:pseudo-linear}
\vv{\Hat{x}} = \mathbf{W(x)} \vv{x}.
\end{equation}
Where do the weights in $\bW$ come from? This is explained in detail in
\cite{milanfar2013tour}, but to summarize, the weights are {\em normalized
affinities} computed between individual or groups of pixels. These affinities
are typically computed using a symmetric positive-definite kernel function
$K_{ij} = k_{ij}(\vx)$. For instance, in the non-local means case
\begin{equation}
k_{ij}(\vx)
= \exp(-\|\vv{R}_i \vv{x} - \vv{R}_j \vv{x}\|^2 /2\gamma^2) = \exp(-\|\vv{R}_{ij}\vv{x}\|^2 /2\gamma^2),
\end{equation}
where $\vv{R}_i$ is a matrix that extracts a pixel or patch centered at a
reference position $i$, and $\gamma$ is a smoothing parameter.
Following~\cite{elmoataz2008}, we call the operator $\bR_{ij}$ the non-local
difference (or discrete derivative) between position $j$ and the reference
position $i$. By definition, the self-weights $k_{ii}(\cdot) = 1$ for all $i$.
The Gaussian non-local means kernel is just one possible choice of kernel, and a
wide variety of others are possible, as noted in~\cite{milanfar2013tour},
illustrated in Table \ref{table:PSD_Kernels}, and elaborated later in this
paper\footnote{Nearly all practical kernels of interest are of the same form,
namely $k_{ij}(\vx) = k(\bR_{ij}\vx)$. Even more specifically, the vast majority
of kernels considered in the image processing community are also isotropic,
namely $k_{ij}(\vx) = k(\|\bR_{ij}\vx\|)$. We will return to this question in
Section \ref{sec:isotropic}.}.

When normalized, these affinities give the weights $W_{ij}$ pointing to each reference position $i$, and summed over the other indices $j$ to produce the output of the (pseudo-linear) filter:
\begin{equation}
\Hat{x}_i = \frac{\sum_j K_{ij} \: x_j}{\sum_{j}K_{ij}} = \sum_j W_{ij} \:x_j.
\end{equation}
Using the matrix notation in (\ref{eq:pseudo-linear}), we rewrite:
\begin{equation}
\vv{\Hat{x}} = \vv{W}(\vv{x}) \vv{x} = \mathbf{D}^{-1}(\vx) \vv{K}(\vv{x})\vv{x}
\end{equation}
where $\bD(\vx) = \text{diag}[d_1,d_2,\cdots,d_N]$ is a diagonal normalization
matrix constructed from the row sums ($d_i = {\sum_{j}K_{ij}}$) of $\bK(\vx)$. In~\cite{romano2016little} we have described in detail some fundamental properties that such smoothing filters satisfy. Of particular interest in the next section is the property that such filters are {\em weakly nonlinear} in the sense that a small perturbation of $\vx$ leaves the filter matrix $\bW(\vx)$ unchanged. As such, their basic behavior strongly mimics certain properties of linear filters, particularly with respect to differentiation; namely, if $f(\vx) =  \bW(\vx)\vx$, then $\nabla f(\vx) = \bW(\vx)$ (see~\cite{romano2016little} for details.)
Correspondingly, it is also shown in \cite{romano2016little} that the (pseudo-quadratic) loss function
\begin{equation}
\phi_\mathit{ker} (\vx) = \frac{1}{2\sigma^2}\:\vx^T(\bI - \bW(\vx)) \vx.
\end{equation}
has gradient directly given by the the pseudo-linear operator
\begin{equation}
\nabla \phi_{ker} (\vx) =  \frac{1}{\sigma^2}(\bI -\bW(\vx)) \vx,
\end{equation}
and Hessian given by
\begin{equation}
\mathcal{H} \phi_{ker} (\vx) =  \frac{1}{\sigma^2}(\bI -\bW(\vx)).
\end{equation}
The operator $\bL(\vx) =\bI -\bW(\vx)$ on the right-hand-side is called the (graph) Laplacian \cite{milanfar2013tour}. Intuitively, if we think of $\bW(\vx)$ as a smoothing (low-pass) filter, then $\bL(\vx)$ is a sharpening (high-pass) filter.  With this correspondence, we can also write the kernel filter in terms of the Laplacian operator as follows:
\begin{equation} \label{eq:quadratic_approx}
\widehat{\vx}_{ker} = \mathbf{W(x)} \mathbf{x} = \vx - \bigl(\bI - \bW(\vx)\bigr) \vx
= \vx - \bL(\vx) \vx,
\end{equation}
It is worth noting that the right-hand side of this equation is in precisely the same form as the {\em shrinkage} $\vv{\Hat{x}} = \vx - \sigma^2 \nabla\phi(\vx)$ equation (\ref{eq:bayes_shrinkage}) we presented earlier in the Bayesian/variational/proximal setting. This observation will be key throughout the rest of the paper.

%-------------------------------------------------------------------------------
\section{Approximation of Regularization by Pseudo-Quadratic Forms}
\label{sec:main_results}

This section contains the main message of this paper, which is to establish a
direct connection between the choice of kernel filter $\vv{W}(\vv{x}) := \bI -
\vv{L}(\vv{x})$ and the penalty function $\phi(\vv{x})$ in the
Bayesian/regularization framework. To this end, the first step is to locally approximate the explicit regularizer with a pseudo-quadratic form of the Laplacian. In particular, for each $\vx$ we set
\begin{equation} \label{eq:pseudo}
\phi(\vx) \approx  \phi_{ker}(\vx) \;\;\;\;\; \implies \;\;\;\; \phi(\vx) =
\frac{1}{2\sigma^2}\:\vx^T \bL(\vx) \vx.
\end{equation}
This is the approximation from which all the rest of our results will flow.
Specifically, given a kernel  $k(\vx)$, we can compute the filter matrix
$\bW(\vx)$ and the Laplacian $\vv{L}(\vv{x}) = \bI - \vv{W}(\vv{x})$, which
immediately gives an approximation to the loss function $\phi(\vx)$. As we will
elaborate below, going in the other direction is much more profitable as it will
help us to localize global approaches. We will dedicate a substantial part of
this paper to working out the details. Before proceeding further, we make some
broad observation that set the stage.
\begin{itemize}
\item The approximation of $\phi$ by $\phi_{ker}$ also implies approximation of the Bayesian denoisers with a kernel filter by way of differentiating (\ref{eq:pseudo}):
\begin{eqnarray}
\label{e:bridge_l}
\nabla \phi(\vx) &=& \frac{1}{\sigma^2} \bL(\vx) \vx \\
\implies \quad \vx - \sigma^2 \nabla \phi(\vx) &=& \bW(\vx)\vx \quad \text{or}
\quad \vv{\Hat{x}}_\mathit{bayes} = \vv{\Hat{x}}_\mathit{ker}.
\end{eqnarray}
We call equation (\ref{e:bridge_l}) the {\em first order} relationship that
bridges the Bayesian/variational framework and the kernel framework. It makes
the surprising and novel observation that very general variational restoration
formulations can be directly associated to filtering by local filters through
the Laplacian operator. Specifically, (\ref{e:bridge_l}) states that the action
of the Laplacian operator on the image approximates the gradient of the implied
regularization function\footnote{Since the (graph) Laplacian is singular with a
zero eigenvalue corresponding to the constant vector $\mathbf{1}$, this means
that corresponding loss must also have a vanishing gradient at $\vx =
\mathbf{1}$.}. Conversely, computing the gradient of a given regularizer can
specify the kernel implicitly in terms of the Laplacian operator $\bL(\vx)$.
Finally, it is worth noting that a special case of this expression also connects
us to ideas in robust statistics \cite{huber1981}, as the gradient $\nabla
\phi(\vx)$ can also be thought of as the {\em influence function} of the
penalty. We shall have much more to say about this later in this section.

\item We can differentiate both sides of (\ref{e:bridge_l}) yet again, yielding the {\em second order} relationship
\begin{equation}
\nabla \left[\nabla \phi(\vx) = \frac{1}{\sigma^2} \mL(\vx) \vx \right] \implies  \mathcal{H}\phi(\vv{x}) = \frac{1}{\sigma^2} \vv{L}(\vv{x}),
\label{e:bridge2}
\end{equation}
where $\mathcal{H}\phi(\vv{x})$ denotes the Hessian. The identity (\ref{e:bridge2}) shows that the loss $\phi(\vx)$ is locally convex if and only if $\bL(\vx)$ (and therefore the corresponding kernel $k(\cdot)$) is positive definite. On the other hand, $\phi(\vx)$ does not have to be globally convex for the relation to hold. Even in the case where $\phi(\vx)$ is non-convex (but twice differentiable), the Hessian still yields a usable kernel (albeit not necessarily a positive definite one).
\end{itemize}

\subsection{First Order Mapping of Losses to Kernels}

Here we will use the explicit first order relationship
\begin{equation}
\shadowbox{$\nabla \phi(\vx) = \frac{1}{\sigma^2} \bL(\vx) \vx$}
\end{equation}
to relate the kernel filters and their corresponding losses. First, let's analyze a couple of relatively simple examples in order to illustrate the above identity.
\paragraph{$\ell_2$ Regularization}
Let $\phi(\vx) = \frac{1}{2}\|\vx\|_2^2$. Therefore, $\nabla \phi(\vx) =
\vx$ (and the Hessian is simply the identity matrix $\mathcal{H}\phi(\vx) =
\bI$). This implies through (\ref{e:bridge_l}) that $\vv{L}(\vv{x}) = \sigma^2
\vv{I}$ and
\begin{equation}
\vv{\Hat{x}}_\mathit{ker} = \bigl(\vv{I} - \vv{L}(\vv{x})\bigr)\vv{x}
= (1 -\sigma^2) \vv{x}.
\end{equation}
The filter is a pointwise multiplicative shrinkage, reminiscent of the James--Stein estimator \cite{james1961}. For comparison, the exact MAP solution is
\begin{equation}
\vv{\Hat{x}}_\mathit{map} = (\vv{I} + \sigma^2
\nabla\phi)^{-1}(\vv{x}) =
(1 + \sigma^2)^{-1} \vv{x},
\end{equation}
which for small $\sigma$ can be approximated using the very same shrinkage expression:
\begin{equation}
\vv{\Hat{x}}_\mathit{map} \approx
(1 - \sigma^2) \vv{x}.
\end{equation}

\paragraph{Dirichlet Energy Regularization} An interesting loss for image processing is the $\ell_2$ norm of the image gradient\footnote{Known also as the Dirichlet energy, and defined in the continuous domain as $\iint \bigl((\partial_x u(x, y))^2 + (\partial_y u(x,y))^2\bigr) \,dx\,dy$, a.k.a.\ the $H^1$ seminorm or $W^{1,2}$ Sobolev seminorm}, which encourages solutions that are smooth. For simplicity, we discuss discretization on a finite one-dimensional grid with periodic boundary conditions.

Given a noisy periodic signal $\vx = \left[ x_0, \ldots, x_{N-1} \right]^T$, the MAP denoiser, regularized by the Dirichlet energy is the minimizer of
\begin{equation}\label{e:h1_minimization}
\vv{\Hat{x}}_\mathit{map} = \operatorname*{argmin}_u
\frac{1}{2\sigma^2} \|\vv{u} - \vv{x}\|^2
+ \phi(\vv{u}), \quad \phi(\vv{u}) = \frac{1}{2}\sum_{n=0}^{N-1} |u_{n+1} - u_n|^2 =\frac{1}{2} \vu^T \mG^T \mG \vu.
\end{equation}
where $\mG$ is the local differencing operator representing circular convolution\footnote{Subscripts are modulo $N$, for example $u_{n+1}$ wraps around to $u_0$ for $n=N-1$.} with the kernel $[1, -1]$.

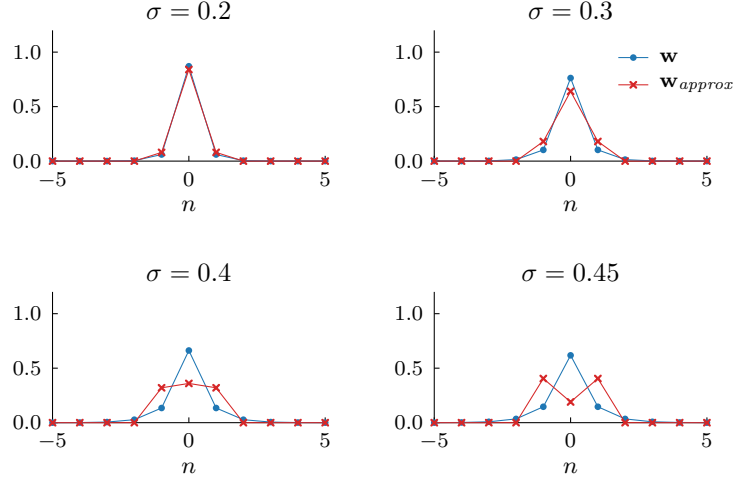
\begin{figure}[!t]
\centering%
\mbox{%
\beginpgfgraphicnamed{figures/h1_filters}%
\input{figures/h1_filters.tikz}%
\endpgfgraphicnamed}
\caption{\label{fig:h1_filters}Comparison of exact (blue) and approximate (red) Dirichlet energy minimizing filter impulse responses.}
\end{figure}

The gradient of $\phi$ is
\begin{equation}
\nabla \phi(\vv{x})_n = -x_{n+1} + 2 x_n - x_{n-1} \quad \implies \quad \nabla \phi(\vv{x}) = \mG^T \mG \vu
\end{equation}
Where we note that $\mL_0 = \mG^T \mG$ is the standard (linear) Laplacian
operator representing convolution with the 3-tap filter $\left[-1, \: 2, \: -1
\right]$. With this, we calculate the approximate MAP solution using (10) as
\begin{equation}
\widehat{\vv{x}}_{approx} =\left(\mI - \sigma^2\mL_0 \right) \vx
\end{equation}
This is to be contrasted against the exact MAP filter which is the following:
\begin{equation}
\widehat{\vv{x}}_{map} =\left(\mI + \sigma^2\mL_0 \right)^{-1} \vx
\end{equation}
The exact and approximate filters share the same eigenvectors, but their
eigenvalues are different\footnote{These eigenvectors are simply the Discrete
Fourier Transform (DFT) basis and the corresponding eigenvalues are the DFT
transformed coefficients.}. Indeed, the exact filter is $(\mI +
\sigma^2\mL_0)^{-1}$ whereas the approximate filter is $\mI - \sigma^2\mL_0$,
and they have eigenvalues respectively of the form $(1+\sigma^2\lambda_i)^{-1}$
and  $1-\sigma^2\lambda_i$, which are approximately equal whenever
$\sigma^2\lambda_i$ is small. We illustrate the $3$-tap approximate denoising
filter and the exact solution in Figure~\ref{fig:h1_filters}, where plots of the
respective filter impulse responses for $N = 256$ for several values of $\sigma$
are shown. Details of the calculation are carried out in the appendix. The
filters are quite close at $\sigma = 0.2$ and become less close for larger
$\sigma$. The approximation breaks down at $\sigma = 1/2\sqrt{2}$, which is a
zero of the approximate filter. For completeness, in
Figure~\ref{fig:h1_distance} in the appendix we also illustrate the $\ell_1$
distance between the filter impulse responses\footnote{This is equivalent to the
$\ell^\infty$ operator distance, which relates the max pointwise error between
$\vv{\Hat{x}}_\mathit{map}$ and $\vv{\Hat{x}}_\mathit{approx}$ as
\begin{equation}
\|\vv{\Hat{x}}_\mathit{map} - \vv{\Hat{x}}_\mathit{approx}\|_\infty = \max_n \Bigl|\sum_{m=0}^{N-1} (w_m - w^\mathit{approx}_m) x_{m-n} \Bigr| \le \|\vv{w} - \vv{w}_\mathit{approx}\|_1 \|\vv{x}\|_\infty.
\end{equation}}.

The above examples illustrated two of the simplest cases where the reduction of
the regularization function to kernel form led to linear operators (shrinkage
and convolution). Next, we consider a much more general class of regularization
functions that lead to more complex (data-dependent) kernels. In particular, we
consider isotropic kernels and their corresponding penalty functions.

\subsubsection{Stationary and Isotropic Losses} \label{sec:isotropic}
In this section we concentrate on special classes of loss functions that allow us to carry out closed-form calculations. In particular, we begin by considering simple losses of the form $\phi(\vx) = \rho(\|\mA\vx\|)$, which depend only on the magnitude of the argument and some linear operator $\mA$. Such loss functions are also sometimes called {\em radial basis} functions, commonly used in approximation theory and machine learning \cite{park91,vert04}. We make the standard regularity assumption on $\rho(\cdot)$ that it is differentiable a.e., and  $\rho(t) \geq 0$, with $\rho(0) = 0$.
%\begin{itemize}
%\item
%\item $\rho(0) = 0$,
%\item $\lim_{t \to 0} \; \frac{\rho'(t)}{t} = 1$
%%\item $\rho(t) = \rho(-t)$
%\end{itemize}

We proceed to calculate the gradient of this loss~\cite{magnus99}:
\begin{eqnarray}
\nabla \phi(\vx) = \nabla \rho(\|\mA\vx\|) = \frac{\rho'(\|\mA\vx\|)}{\|\mA\vx\|}\mA^T\mA \vx
\end{eqnarray}
Invoking (\ref{e:bridge_l}), for all $\vx$ we have\footnote{If $\mA$ is any orthogonal matrix, then $\mA^T \mA = \mI$ and $\|\mA\vx\| = \|\vx\|$, which means that $\frac{\rho'(\|\vx\|)}{\|\vx\|}$ are effectively assigning the eigenvalues of the Laplacian matrix $\mL(\vx)$. The problem of finding the kernel in this case can be viewed as an {\em inverse eigenvalue} problem of solving for a positive semi-definite matrix with given eigenvalues.}
\begin{eqnarray}
\frac{1}{\sigma^2} \mL(\vx) \vx = \frac{\rho'(\|\mA\vx\|)}{\|\mA\vx\|}\mA^T\mA \vx
\end{eqnarray}
This implies that
\begin{eqnarray}
\left( \frac{1}{\sigma^2} \mL(\vx) -
\frac{\rho'(\|\mA\vx\|)}{\|\mA\vx\|}\mA^T\mA \right) \vx = \mathbf{0},
\end{eqnarray}
or
\begin{eqnarray} \label{eq:kernelLaplacianApprox}
 \frac{1}{\sigma^2} \mL(\vx) = \frac{\rho'(\|\mA\vx\|)}{\|\mA\vx\|}\mA^T\mA.
\end{eqnarray}

Next, we generalize the loss functions to the following form:
\begin{equation} \label{eq:penalty_isotropic}
\phi(\vx) = \sum_{i, j} h_{ij} \rho (\|\bR_{ij}\vx\|)
\end{equation}
where $h_{ij} = h_{i-j}$ is a positive stationary but otherwise arbitrary weight
functions. In this form, the class of functions we consider includes many
well-known losses such as those shown in Table \ref{table:PSD_Kernels}.
With this definition, the gradient of $\phi(\cdot)$ is:
\begin{eqnarray}
\nabla \phi(\vx) =\sum_{i, j} h_{ij} \nabla \rho(\|\bR_{ij}\vx\|)
\end{eqnarray}
where
\begin{eqnarray} \label{eq:gradRho}
\nabla \rho(\|\bR_{ij}\vx\|) & = & \frac{\rho'(\|\bR_{ij}\vx\|)}{\|\bR_{ij}\vx\|}\bR_{ij}^T\bR_{ij} \:\vx
\end{eqnarray}
As we will show in detail shortly, this class of losses will induce stationary and isotropic kernels that depend strictly on the norm (and not the direction) of the (non-local) difference: $K_{ij} = k(\|\bR_{ij}\vx\|)$. This is not a severe restriction for our purposes, and is routine for many of the kernels used in image processing, such as those illustrated in Table \ref{table:PSD_Kernels}.  Isotropic kernels are also well-behaved, and convenient tools for their analysis and computation exist. Bochner's Theorem \cite{rahimi07}, for instance, gives them a useful spectral
representation: namely, isotropic positive-definite kernels are the Fourier transform of a finite non-negative-valued function (measure)~\cite{rahimi07}. Of course, one can construct new kernels by adding, multiplying, composing, etc. as is well-known in the machine learning community \cite{genton01}.

Now, let's take steps to extract the implied kernels. In analogy to (\ref{eq:kernelLaplacianApprox}), we have
\begin{eqnarray} \label{eq:sumofinfluence}
\frac{1}{\sigma^2} \bL(\vx) =  \sum_{ij} h_{ij} \frac{\rho'(\|\bR_{ij}\vx\|)}{\|\bR_{ij}\vx\|}\bR_{ij}^T\bR_{ij}
\end{eqnarray}
To clarify what the RHS really means, we remind the careful reader that $\bR_{ij} = \bR_{i} - \bR_{j}$. Therefore,
\begin{eqnarray} \label{eq:theRs}
\bR_{ij}^T\bR_{ij} & = & (\bR_{i} - \bR_{j})^T(\bR_{i} - \bR_{j}) \\
& = & \bR_{i}^T\bR_i + \bR_{j}^T\bR_{j} - \bR_{i}^T\bR_{j} -\bR_{j}^T\bR_{i}.
\end{eqnarray}
To simplify further, we pause to interpret each of the terms in the above. $\bR_{i}$ is an operator that {\em extracts} a pixel or a (symmetric) patch centered at the position $i$. The transpose of this operator $\bR_{i}^T$ takes a given patch or pixel and {\em places} it at position $i$ in an image of all zeros. Therefore, $\bR_{i}^T\bR_{i}$ (and $\bR_{j}^T\bR_{j}$) does nothing more than extract a patch and put it back where it came from; thereby only contributing a value of $1$ to the relevant {\em diagonal} element of the matrix\footnote{When such terms are summed, the resulting patches placed in the image must be summed at locations where they overlap. If $\mR$ select only one pixel in the simplest case, then there is no overlap.} $\bR_{ij}^T\bR_{ij}$. Correspondingly, $\bR_{j}^T\bR_{i}$ picks off a patch at position $i$ and places it at position $j$. In matrix terms, $\bR_{j}^T\bR_{i}$ contributes a value of $1$ to the corresponding {\em off-diagonal} position each time pixels from the patches $i$ and $j$ overlap\footnote{Since the left-hand-side of (\ref{eq:theRs}) is symmetric by definition, we also have that $\bR_{j}^T\bR_{i} = \bR_{i}^T\bR_{j}$.}. Together, these observations help us write the RHS of (\ref{eq:sumofinfluence}) as a sum of two sets of terms: one set for the diagonal elements and another for the off-diagonals, as follows:
\begin{eqnarray}
\sum_{ij} h_{ij} \frac{\rho'(\|\bR_{ij}\vx\|)}{\|\bR_{ij}\vx\|}\bR_{ij}^T\bR_{ij} & = &  2\sum_{i}h_{ii}\frac{\rho'(\|\bR_{ii}\vx\|)}{\|\bR_{ii}\vx\|}\bR_{i}^T\bR_{i}- 2\sum_{i\neq j}h_{ij}\frac{\rho'(\|\bR_{ij}\vx\|)}{\|\bR_{ij}\vx\|}\bR_{i}^T\bR_{j}
%\\
%& = & 2\sum_{i}h_{ii}\frac{\rho'(\|0\|)}{\|0\|}\bR_{i}^T\bR_{i}- 2\sum_{i\neq j}h_{ij}\frac{\rho'(\|\bR_{ij}\vx\|)}{\|\bR_{ij}\vx\|}\bR_{i}^T\bR_{j}  \\
 %& = & 2\sum_{i}h_{ii}\bR_{i}^T\bR_{i}- 2\sum_{i\neq j}h_{ij}\frac{\rho'(\|\bR_{ij}\vx\|)}{\|\bR_{ij}\vx\|}\bR_{i}^T\bR_{j}
\end{eqnarray}
%Therefore
%\begin{eqnarray} \label{eq:solveforL}
%\frac{1}{2\sigma^2} \bL(\vx)  =  \sum_{i}h_{ii}\bR_{i}^T\bR_{i} - \sum_{i\neq j}h_{ij}\frac{\rho'(\|\bR_{ij}\vx\|)}{\|\bR_{ij}\vx\|}\bR_{i}^T\bR_{j}.
%\end{eqnarray}
Here we have an expression where the first term on the RHS fixes the diagonal elements of $\bL(\vx)$, and the second term on the RHS determines its off-diagonal elements. Our hope is that this expression can be used to solve for the elements of the kernel matrix $\bK_{ij}(\vx)$.

To do this, let's recall the definition of the Laplacian operator once again:
\begin{equation}
\bL(\vx) = \bI - \bW(\vx) = \bI - \bD^{-1}(\vx)\bK(\vx)
\end{equation}
In the context of solving for a specific kernel $\bK(\vx)$, the presence of the term $\bD^{-1}(\vx)$ performing pixel-wise division complicates matters significantly. First, the pixel-wise division is a computational complexity we would rather avoid; second, the division prevents us from solving directly for the kernels of interest; and third, the division renders the filter and the Laplacian matrix non-symmetric. Fortunately, all of these issues can be overcome by a nice computational simplification, whereby we can eliminate the per pixel division~\cite{milanfar2016new}. We explain this idea next.

\begin{table}
\centering
\begin{tabular}{l@{\hspace{2em}}l@{\hspace{2em}}l}
\hline
Name & Kernel & Penalty \\
\hline
Quadratic  & $\exp(-\|\vx_i-\vx_j\|^2 /s^2) $ &
$\frac{s^2}{4} \bigl(1 - \exp(-\|\vx_i-\vx_j\|^2 / s^2)\bigr)$  \\
Exponential & $\exp(-\|\vx_i-\vx_j\|_1/s) $ &
$\frac{s}{2}(1 - \exp(-\|\vx_i-\vx_j\|_1/s)$ \\
Cauchy & $ \left(1+\frac{\|\vx_i-\vx_j\|^2}{s^2} \right)^{-1} $ &
$\frac{s^2}{4} \log(1 + \|\vx_1-\vx_j\|^2/s^2)$ \\
Dirichlet & constant &  $\|\vx_i-\vx_j\|^2 $ \\
Total Variation & $2/\|\vx_i-\vx_j\|_1$ &  $\|\vx_i-\vx_j\|_1 $ \\
Lorentzian & $4/(1 + \|\vx_i-\vx_j\|^2)$ & $ \log(1+\|\vx_i-\vx_j\|^2) $ \\
\hline
\end{tabular}
\caption{Some well-known isotropic, positive-definite kernels $k(\|\bR_{ij}\vx\|)$
and corresponding penalities $\phi(\|\bR_{ij}\vx\|)$.}
\label{table:PSD_Kernels}
\end{table}

\subsubsection{Avoiding pixel-wise normalization}
The {\em normalized} Laplacian matrix $\bL(\vx) =
\mathbf{D}^{-1}(\vx) \mathbf{K(x)}$ can be written in terms of the {\em unnormalized} Laplacian matrix as follows:
\begin{equation}
\bL(\vx) \approx \alpha \left(\bD(\vx) - \bK(\vx) \right),
\end{equation}
where $\alpha$ is a fixed parameter. As shown in \cite{milanfar2016new}, for practical kernels this approximation tends to be quite accurate. To illustrate the approximation, consider the optimization problem
\begin{equation}
\min_{\alpha} \| \bL (\vx)- \alpha \left(\bD(\vx) - \bK(\vx) \right) \|_F^2
\end{equation}
where the norm is the Frobenius norm. The objective to be minimized is quadratic
in $\alpha$. Therefore, upon differentiating and setting to zero, we are led to the global
minimum solution :
\begin{equation} \label{roots}
\alpha = \frac{\tr(\bK\bD^{-1}\bK) -2\tr(\bK) +\tr(\bD)}{\tr(\bK^2) -2\tr(\bK \bD) +\tr(\bD^2)}
\end{equation}
When $N$ is large enough, the right-hand-side is closely approximated by\footnote{Each $d_i$ is individually the sum of $N$ terms bounded by one, so $d_i
=\mathcal{O}(N)$ and  $\sum_i d_i = \mathcal{O}(N^2)$.
So more to the point, a small enough $\alpha \propto \mathcal{O}(N^{-1})$ will give an adequate
approximation as illustrated in detail in~\cite{milanfar2016new}.}
\begin{equation}
\alpha = \frac{1}{\mbox{mean}(d_i)} = \frac{N}{\sum_i d_i}.
\end{equation}

Now we have all the ingredients in place to solve for an underlying kernel. With the unnormalized Laplacian, we write
\begin{eqnarray} \label{eq:solveforK}
\frac{\alpha}{2\sigma^2} \left(\bD(\vx) - \bK(\vx) \right)   =  \sum_{i}h_{ii}\bR_{i}^T\bR_{i} - \sum_{i\neq j}h_{ij}\frac{\rho'(\|\bR_{ij}\vx\|)}{\|\bR_{ij}\vx\|}\bR_{i}^T\bR_{j}.
\end{eqnarray}
Equating the elements of the LHS to the respective elements of the RHS we have
\begin{eqnarray}
\frac{\alpha}{2\sigma^2}(d_i-1) & = & h_{ii}, \\
\frac{\alpha}{2\sigma^2}k(\|\bR_{ij}\vx\|) & = & h_{ij}
\frac{\rho'(\|\bR_{ij}\vx\|)}{\|\bR_{ij}\vx\|}.
\end{eqnarray}
To summarize,
\begin{eqnarray}
\phi(\vx) = \sum_{i, j} h_{ij} \rho (\|\bR_{ij}\vx\|)\:\:\:\:\: \implies
\:\:\:\:\: k(\|\bR_{ij}\vx\|) & = & \frac{2\sigma^2}{\alpha} h_{ij}
\frac{\rho'(\|\bR_{ij}\vx\|)}{\|\bR_{ij}\vx\|}.
\end{eqnarray}
%Or to say this more compactly we can define the variable $t = \|\bR_{ij}\vx\|$:
%\begin{equation}
%k(t) \propto \frac{\rho'(t)}{t}
%\end{equation}
This interesting expression directly relates the kernel to the loss function. Let's simplify notation by defining the index $l = |i-j|$, and the variable $t_l = \|\bR_{ij}\vx\|$:
\begin{equation}
k(t_l) =\frac{2\sigma^2h_l}{\alpha} \frac{\rho'(t_l)}{t_l}.
\end{equation}
The RHS in particular is noteworthy, as it is exactly the expression for the
weights seen in the reweighted least squares formulation of M-estimators in
robust statistics \cite{huber1981}. The above equation is a special
case of the first-order kernel relationship (\ref{e:bridge_l}) for scalar,
isotropic, shift-invariant kernels: in this case, it is possible to
divide both sides by $\vv{x}$ (above, $t_l$). Division isn't possible in
(\ref{e:bridge_l}) (where $\vv{x}$ is a vector), which is a much more general
formulation and puts these ideas in a broader context.

The regularity assumptions we made earlier
on $\rho$ ensure that the implied kernel $k$ is (a.e.) well defined and
differentiable. Next, we examine some specific pairs of kernels and
corresponding regularization functions. To simplify the exposition, we set $h_l=
\delta(l)$, the Dirac delta function, and therefore drop the dependence on the
index $l$. This simplified form is
 \begin{equation} \label{eq:simplek}
k(t)  =\frac{2\sigma^2}{\alpha} \frac{\rho'(t)}{t}  \quad \iff \quad \rho(t) =
\frac{\alpha}{2\sigma^2}\int \tau k(\tau) d\tau + c,
\end{equation}
where due to symmetry, it is sufficient to consider the case of $t>0$. We compute some specific examples in detail below.

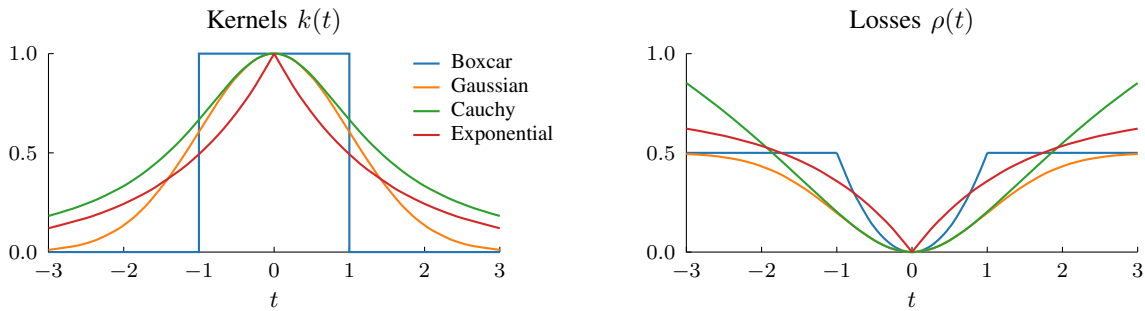
\begin{figure}
\centering%
\begin{subfigure}[t]{0.45\linewidth}%
\centering%
\beginpgfgraphicnamed{figures/kernels}%
\input{figures/kernels.tikz}%
\endpgfgraphicnamed
\end{subfigure}%
\begin{subfigure}[t]{0.45\linewidth}%
\centering%
\beginpgfgraphicnamed{figures/losses}%
\input{figures/losses.tikz}%
\endpgfgraphicnamed%
\end{subfigure}%
\caption{Examples of kernels $k(t)$ and associated losses $\rho(t)$. Left:
several common kernel functions $k(t)$. Right: corresponding
loss functions $\rho(t)$ through first-order approximation for
$\gamma = 1$, $\sigma = 1$. }
\label{fig:typical_kernels}
\end{figure}

\textbf{Examples of kernels $k(t)$ to losses $\rho(t)$} First,  consider some familiar kernels $k(t)$, with $\gamma$ denoting generically their smoothing parameter. We will illustrate how these kernels map to corresponding loss functions $\rho(t)$.  Plots of the functions are shown in Figure~\ref{fig:typical_kernels}.

\subsubsection*{Boxcar Kernel} This kernel gives equal weights to all samples within a distance $\gamma$
\begin{align*}
k(t) &=
\begin{cases}
1 & \text{if}\  t  \le \gamma, \\
0 & \text{otherwise},
\end{cases} \\
\rho (t) & =\frac{\alpha}{2\sigma^2}
\begin{cases}
 t^2 & \text{if}\  t \le \gamma, \\
\gamma^2 & \text{otherwise}.
\end{cases}
\end{align*}
This implied loss function is simply a ``clipped'' version of the standard $\ell_2$ loss, which naturally makes it more robust to outliers.

\subsubsection*{Gaussian Kernel} This kernel is the most commonly used, for
instance in the context of non-local means and bilateral filters. The resulting loss function is known as the Welsch loss~\cite{welsch1977},  and behaves like $\ell_2$ near zero and for large $t$ asymptotically clips to the same constant value as we derived above for the boxcar. It is interesting, and consistent with what has been observed elsewhere \cite{black1998robust,elad2002origin}, that the use of even the most common kernel (Gaussian) in the context of locally-adaptive filtering results in robust behavior in the implied loss.
\begin{align*}
k(t) &= e^{- \frac{t^2}{2 \gamma^2}}, \\
\rho (t) &= \frac{\alpha\gamma^2}{2\sigma^2} \left( 1- e^{- \frac{t^2}{2 \gamma^2}} \right).
\end{align*}

\subsubsection*{Cauchy Kernel} This kernel is uncommon in image processing. The
resulting loss function is sometimes called the Lorentzian. The implied loss
function (again) behaves like $\ell_2$ near the origin, but grows
logarithmically for large $t$.
\begin{align*}
k(t) &= \frac{1}{1 + \frac{t^2}{2 \gamma^2}}, \\
\rho(t) &= \frac{\alpha\gamma^2}{2\sigma^2}  \log\left(1+ \frac{t^2}{2\gamma^2} \right).
\end{align*}
\subsubsection*{Exponential Kernel} This kernel is the only one that results in
a loss function that behaves qualitatively different than the others near $t=0$.
Namely, it behaves like $\ell_1$ for small $t$ and settles to a constant for
large $t$
\begin{align*}
k(t) &= e^{- \frac{|t|}{\sqrt{2} \gamma}}, \\
\rho(t) &= \frac{\sqrt{2}\alpha\gamma}{2\sigma^2}(1- e^{-\frac{|t|}{\sqrt{2}
\gamma}} ).
\end{align*}

\textbf{Examples of losses $\rho(t)$ to kernels $k(t)$: } Now let's go in the
other direction, where we specify some well-known loss functions $\rho(t)$, and
compute the implied kernels. Since any constant scaling of these kernels is
canceled after normalization and therefore unimportant, we shall drop them.
Plots of several of these functions  are shown as part of a much larger family
in Figure~\ref{fig:robust_loss}.

\subsubsection*{Dirichlet Energy} The standard $\ell_2$ loss applied to the
difference operator $t = \|\bR_{ij}\vx\|$ as $\rho(t) = \frac{1}{2}t^2$ is the
Dirichlet energy defined earlier\footnote{It is important to note that this loss
should not be confused with $\|\vx\|^2$ which is the $\ell_2$ loss applied
directly to the image.}.  The result is a constant (averaging) kernel
\begin{equation}
k(t) \propto \frac{t}{t}= \text{constant},
\end{equation}
which is consistent with the approximate MAP solution, and when modulated by the weights $h_{l}$ gives a local spatial convolution kernel as described in the appendix.

\subsubsection*{Total Variation} This penalty is $\rho(t) = t$ (with identical weights $h_l = 1/N$), and yields
\begin{equation}
k(t) \propto \frac{1}{t}.
\end{equation}
As is apparent, the resulting kernel gives huge weight to pixels (patches) that are very similar, and hence acts largely locally. This means that the resulting filter tends to behave mostly like a linear averaging filter in low contrast areas, while respecting edges and therefore producing piecewise constant results. The TV penalty as a local filter has been studied in \cite{louchet2011total} and our observation is consistent with the conclusions in that work.

\subsubsection*{Huber Loss} The Huber loss is meant to strike a balance between the use of the $2$-norm near the origin and the $1$-norm away from the origin. This is a very useful loss function that yields appropriate behavior for both high and low signal-to-noise regions. For instance, when used in the context of denoising, it tends to aggressively smooth areas that are ``flat'' (i.e. very similar) and be less aggressive in smoothing areas that contain discontinuities. The resulting kernel reflects exactly this behavior: It is constant for small values of $t$, and decays in inverse proportion to increasingly large values of $t$.
\begin{equation}\label{e:huber}
\rho (t) =
\begin{cases}
\frac{1}{2} t^2 & \text{if}\ | t | \le \gamma, \\
\gamma |t| - \frac{1}{2} \gamma^2 & \text{otherwise},
\end{cases}
\end{equation}

\begin{align*}
k(t) &\propto
\begin{cases}
 1 & \text{if}\ t  \le \gamma, \\
\frac{\gamma}{|t|}  & \text{otherwise}.
\end{cases}
\end{align*}

\begin{figure}
\centering%
\begin{subfigure}[t]{0.45\linewidth}%
\centering%
\beginpgfgraphicnamed{figures/kernels_barron}%
\input{figures/kernels_barron.tikz}%
\endpgfgraphicnamed%
\label{fig:kernel_general}
\end{subfigure}
% \begin{subfigure}[t]{0.4\linewidth}
% \centering
% \includegraphics[width=\linewidth]{figures/activation2}
% \label{fig:activation}
% \end{subfigure}
% \newline
\begin{subfigure}[t]{0.45\linewidth}%
\centering%
\beginpgfgraphicnamed{figures/losses_barron}%
\input{figures/losses_barron.tikz}%
\endpgfgraphicnamed%
\label{fig:regularization_general}
\end{subfigure}
\caption{Kernel functions $k(t)$ and
their loss functions $\rho(t)$ for the general robust loss function,
presented by Barron~\cite{barron2017more}, sweeping over $\beta$, and fixing $c
= 1$. }
\label{fig:robust_loss}
\end{figure}
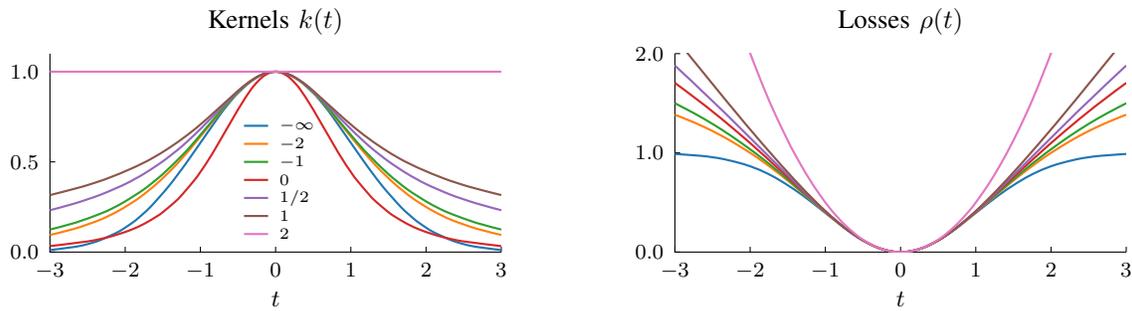

For the sake of completeness, let's also carry out the computations for a wide
class of loss functions that span a gamut of behaviors including the $\ell_2$
and $\ell_1$. For this purpose, we consider the family described
in~\cite{barron2017more}, which includes the Welsch and Lorentzian losses
discussed earlier,
\begin{eqnarray} \label{eq:robust_loss}
\rho (t) &=
\begin{cases}
\log \left(\frac{1}{2} \left(\frac{t}{\gamma} \right)^2 + 1\right) & \text{if}\
\beta = 0, \\
1 - e^{-\frac{1}{2} \left( \frac{t}{\gamma} \right)^2} & \text{if}\  \beta =
-\infty, \\
\frac{z(\beta)}{\beta} \left( \left(
\frac{\left(\frac{t}{\gamma}\right)^2}{z(\beta)} + 1 \right)^{\beta / 2} - 1
\right) & \text{otherwise},
\end{cases}
\end{eqnarray}
where $\beta$, and $\gamma$ are order and strength parameters, and $z(\beta) =
\text{max}(1, 2 - \beta)$. Plots of the functions are shown in
Figure~\ref{fig:robust_loss}. Their corresponding kernel functions are
\begin{align*}
k(t) &\propto
\begin{cases}
\left( \frac{1}{2} \left(\frac{t}{\gamma}\right)^2 + 1\right)^{-1} & \text{if}\
\beta = 0, \\
 e^{-\frac{1}{2} \left( \frac{t}{\gamma} \right)^2} & \text{if}\  \beta =
-\infty, \\
\left( \frac{\left(\frac{t}{\gamma}\right)^2}{z(\beta)} + 1 \right)^{\beta / 2 - 1} &
\text{otherwise}.
\end{cases}
\end{align*}

%-------------------------------------------------------------------------------
\section{Second Order Correspondence of Kernels and Regularizers}
\label{sec:main_results2}

In the previous section we illustrated the first order approximation of regularizers with kernel filters. This relationship allowed us to directly compute kernels given the Bayesian/variational/proximal description. These convenient formulas were derived for isotropic losses $\rho(t)$ which simplified the calculations to yield locally adaptive kernels. In this section, we ask the question of whether more general expressions exist that apply to the non-isotropic cases. For this purpose, we have to resort to second order analysis. Ironically, the second order analysis, by virtue of being less accurate is less generally useful in practice. But it does have the advantage that it illuminates the geometry of the problem at hand. Namely as we shall see below, the second order analysis relates the {\em curvature} of the loss function $\phi(\vx)$ to the Laplacian operator. Given $\phi(\vx)$, we can compute the Hessian $\mathcal{H}\phi(\vv{x})$ (assuming it exists) and directly obtain $\vv{L}(\vv{x})$:
\begin{equation}
\shadowbox{$\mathcal{H}\phi(\vv{x}) = \frac{1}{\sigma^2}\vv{L}(\vv{x})$}
\end{equation}
which then implies the filter $\bW(\vx) = \mI - \mL(\vx)$. If the level sets of the loss $\phi(\vx) - c = 0$ describe a smooth hypersurface\footnote{A hypersurface (such as a hypersphere) is defined by a single implicit equation, at least locally.} of dimension $N-1$ embedded in $\mathbb{R}^N$, then the Hessian $\mathcal{H}\phi(\vv{x})$ measures the local curvature of the graph of this function. In particular, the Hessian is the second fundamental form of the surface, which contains as its eigenvectors the principal curvature directions \cite{spivak1970comprehensive}. On the right-hand side we have the Laplacian {\em operator}. This indicates that with the second order approximation, the Laplacian operator (and therefore the filter weights) are assigned by the curvature of the loss function -- an interesting and reasonable intuition.

What guarantee do we have that the Hessian even exists? Interestingly, for convex $\phi(\vv{x})$, a classical result due to Alexandrov~\cite{alexandrov39} states that $\nabla \phi(\vx)$ exists and is
differentiable almost everywhere (a.e.); therefore the Hessian is well-defined
a.e. and we are in business. In particular, let's again employ the {\em unnormalized} Laplacian to write
\begin{equation}
\mathcal{H}\phi(\vv{x}) = \frac{\alpha}{\sigma^2}(\bD(\vx) - \bK(\vx)).
\end{equation}
If we write this element-wise, the diagonal and off-diagonal elements of the
kernel weight matrix are the solutions of the equations
\begin{equation}
\frac{\partial^2 \phi(\mathbf{x})}{\partial x_i^2} =  \frac{\alpha}{\sigma^2} \sum_{j \ne i} k_{ij}(\vx) \quad \text{and} \quad \frac{\partial^2 \phi(\mathbf{x})}{\partial x_i \partial x_j} = -\frac{\alpha}{\sigma^2} \;k_{ij}(\mathbf{x}).
\end{equation}
These equations can directly specify the kernel in terms of mixed partial derivatives of the loss function.

\textbf{Example 1}: The standard linear-quadratic loss is a useful and simple case:
\begin{equation}
\phi(\vx) = \vx^T \bA \vx + \bb^T\vx + c \quad \implies \quad \mathcal{H}\phi(\vx) = \bA \quad \implies \quad \mK(\vx) \propto \mI - \mA
\end{equation}
where $\bA$ is assumed symmetric and positive semi-definite. So any (linear-) quadratic loss function yields a (constant) linear kernel filter.

\textbf{Example 2}:
What about the opposite direction? Namely, given a symmetric kernel
$k(\cdot,\cdot)$, and the corresponding symmetric Laplacian $\vv{L}(\vv{x})$,
can we try to solve for a $\phi(\vx)$? This is a far harder question as it
implies a nonlinear elliptic partial differential equation for which in general
a solution may not even exists. Indeed, not every vector field $f:\mathbb{R}^N
\rightarrow \mathbb{R}^N$ corresponds to the gradient of some scalar function $f
\stackrel{?}{=} \nabla \phi$, and similarly, an arbitrary $\vv{L}(\vv{x})$ might
not be a valid Hessian. This equation falls under the class of Monge--Amp\`ere
equations~\cite{gutierrez2016monge}, which model (among other things) the
relationship between the curvature of hypersurfaces and their explicit
descriptions in Euclidean coordinates. Finding solutions is challenging,
especially in high dimensions. Whether a solution exists for general cases is an
open question.

\subsection{Specializing to isotropic losses, and comparison to first order} As we did earlier in the first order analysis, consider the particular case of (a.e. twice differentiable) losses of the form $\phi(\vx) = \rho(\|\mA\vx\|)$. This gives
\begin{equation} \label{eq:second_order_expression}
\mathcal{H}\phi(\vv{x}) = \frac{\rho'(\|\mA\vx\|)}{\|\mA\vx\|} \mA^T \mA +
 \left( \rho''(\|\mA\vx\|) - \frac{\rho'(\|\mA\vx\|)}{\|\mA\vx\|} \right) \mA^T \bu \bu^T \mA
\end{equation}
where $\bu = \frac{\mA\vx}{\|\mA\vx\|}$. It is interesting to note that the first (order) term on the right-hand-side is identical to the expression we derived earlier in (\ref{eq:kernelLaplacianApprox}). Let's rewrite (\ref{eq:second_order_expression}):
\begin{equation}
\mathcal{H}\phi(\vv{x}) = \frac{\rho'(\|\mA\vx\|)}{\|\mA\vx\|} \mA^T\left( \mI -  \bu \bu^T  \right)\mA +
\rho''(\|\mA\vx\|) \mA^T \bu \bu^T \mA.
\end{equation}
The first term in this expression is in fact singular for all $\vx$!
\begin{eqnarray}
\frac{\rho'(\|\mA\vx\|)}{\|\mA\vx\|} \mA^T\left( \mI -  \bu \bu^T \right) \mA \vx & = & \frac{\rho'(\|\mA\vx\|)}{\|\mA\vx\|} \mA^T\left( \mI -  \frac{\mA\vx\vx^T\mA^T}{\|\mA\vx\|^2} \right) \mA \vx \\
& = & \frac{\rho'(\|\mA\vx\|)}{\|\mA\vx\|} \mA^T\left( \mA\vx -  \frac{\vx^T\mA^T\mA\vx}{\|\mA\vx\|^2} \mA \vx \right)  \\
& = & \frac{\rho'(\|\mA\vx\|)}{\|\mA\vx\|} \mA^T\left( \mA\vx - \mA \vx \right) = \mathbf{0}
\end{eqnarray}
Therefore, the expression for the Hessian simplifies considerably as we consider the action of this operator on $\vx$
\begin{eqnarray}
\mathcal{H}\phi(\vv{x}) \: \vx & = & \frac{\rho'(\|\mA\vx\|)}{\|\mA\vx\|} \mA^T\left( \mI -  \bu \bu^T  \right)\mA \vx +
\rho''(\|\mA\vx\|) \mA^T \bu \bu^T \mA \vx  \\
& = & \mathbf{0} + \rho''(\|\mA\vx\|) \mA^T \bu \bu^T \mA \vx \\
& = & \rho''(\|\mA\vx\|) \mA^T \mA \vx. \label{e:phihessian}
\end{eqnarray}
Hence, the second order analysis gives the Hessian as
\begin{equation}
\mathcal{H}\phi(\vv{x}) =  \rho''(\|\mA\vx\|) \mA^T \mA.
\end{equation}
To summarize, for losses of the form
\begin{eqnarray}
\phi(\vx) = \rho (\|\mA\vx\|)
\end{eqnarray}
we have:
\begin{eqnarray}
\text{First-order kernel approximation:} \quad & \quad \mL(\vx)  = & \sigma^2 \frac{\rho'(\|\mA \vx\|)}{\|\mA\vx\|} \mA^T \mA\\
\text{Second-order kernel approximation:} \quad & \quad \mL(\vx)  = & \sigma^2 \rho''(\|\mA\vx\|)  \mA^T \mA
\end{eqnarray}
Examples of how well the first and second order solutions work are presented in the next section. As seen in Figs. $5-9$ of Section \ref{sec:experiments}, both of these approximations are accurate for small $\sigma$, though not surprisingly, the second order approximation is consistently less accurate. The reader might wonder under what circumstances the two approximations might coincide. To address this questions, let's again denote $t = \|\mA\vx\|$. It is obviously the case that whenever $\rho(t) \approx t^2$ locally, the two approximations coincide because $\rho''(t) = \rho'(t)/t$. As we've noted earlier, this is the case for many losses of interest, including the Huber loss and the wider class of robust losses in (\ref{eq:robust_loss}), all of which behave quadratically for small $t \leq \gamma$.

Another interesting scenario is robust losses that are {\em redescending}
\cite{huber1981,welsch1977}. Redescending losses are those where, for large $t >
\gamma$, the second derivative $\rho''(t)$ is small\footnote{specifically, it is
bounded~\cite{barron2017more} by $1/\gamma^2$.}. Therefore, the Hessian of the
redescending loss for large $t = \|\mA\vx\|$, when applied to any $\vx$ is
\begin{eqnarray}
\mathcal{H}\phi(\vv{x}) \vx & \approx & \left(\frac{\rho'(\|\mA\vx\|)}{\|\mA\vx\|} \mA^T \mA - \frac{\rho'(\|\mA\vx\|)}{\|\mA\vx\|} \mA^T \bu \bu^T \mA\right) \vx  \\
 & = & \frac{\rho'(\|\mA\vx\|)}{\|\mA\vx\|} \mA^T\left( \mI -  \bu \bu^T \right)
\mA \vx = \mathbf{0}.
\end{eqnarray}
That is, the Hessian in the large argument regime is in fact singular. In retrospect, this should not be entirely surprising given that we would expect redescending losses to censor vectors that have large magnitude $\|\mA\vx\|$.
As before, we can specialize the above observations to the case where the loss is a sum of radial basis functions acting on patches of the image as follows:
\begin{eqnarray}
\phi(\vx) = \sum_{i, j} h_{ij} \rho (\|\bR_{ij}\vx\|)
\end{eqnarray}
which again, in summary, yields the corresponding kernel functions:
\begin{eqnarray}
\text{First-order kernel approximation:} \quad & \quad k(\|\bR_{ij}\vx\|)  = &\frac{2\sigma^2}{\alpha}h_{ij}  \frac{\rho'(\|\bR_{ij}\vx\|)}{\|\mR_{ij}\vx\|} \\
\text{Second-order kernel approximation:} \quad & \quad k(\|\bR_{ij}\vx\|)  = & \frac{2\sigma^2}{\alpha} h_{ij} \rho''(\|\bR_{ij}\vx\|)
\end{eqnarray}

\section{Experimental Results}
\label{sec:experiments}

In this section we demonstrate the use of the results derived in the earlier
sections.

\subsection{Approximating Total Variation Regularized Denoising}

In the loss-to-kernel direction, we consider a form of total variation (TV)
denoising
\begin{equation}\label{e:tvmap}
\operatorname*{arg\,min}_u \,
\frac{1}{2\sigma^2} \|\vv{u} - \vv{x}\|_2^2 + \phi(\vv{u}),
\quad \phi(\vv{u}) = \frac{1}{2} \sum_i \sum_j
h_{ij} \rho(\lvert x_i - x_j \rvert)
\end{equation}
where $\rho$ is the Huber function (\ref{e:huber}) with $\gamma=5$ and
$h_{ij}$ is one in an $11\times 11$ spatial neighborhood for $(i-j) \in
\{5,\ldots,5\}\times\{-5,\ldots,5\}$ and zero otherwise. We include a
$1/2$ factor since the sum in $\phi$ counts each difference $|x_i -
x_j|$ twice. We approximate the solution of the MAP problem (\ref{e:tvmap}) with
kernelized filters using the first- and second-order kernel relationships.

Through the first-order relationship (\ref{e:bridge_l}), we obtain the kernel
filter
\begin{align}
\Hat{\vv{x}}^\textit{first}
&= \big(\vv{I} - \sigma^2
\nabla\phi(\vv{x})\bigr) \vv{x}
= \bigl(\vv{I} - \tfrac{1}{2} \sigma^2
{\textstyle\sum_i\sum_j h_{ij}
\frac{\rho'(|\vv{R}_{ij}\vv{x}|)}{|\vv{R}_{ij}\vv{x}|} \vv{R}_{ij}^T \vv{R}_{ij}
}\bigr)
\vv{x}
\end{align}
where $\vv{R}_{ij}$ is the matrix $(\vv{R}_i - \vv{R}_j)$ such that $\vv{R}_{ij}
\vv{x} := x_i - x_j$.
The roles of $i$ and $j$ are symmetric in $h_{ij}
\frac{\rho'(|\vv{R}_{ij}\vv{x}|)}{|\vv{R}_{ij}\vv{x}|}$, and the expression
$\vv{R}_{ij}^T \vv{R}_{ij} \vv{x} =
\vv{R}_i^T (x_i - x_j) + \vv{R}_j^T (x_j - x_i)$ contributes $(x_i - x_j)$ to output pixel
$i$ and $(x_j - x_i)$ to pixel $j$. So for a given output pixel $k$,
\begin{align}
\Hat{x}^\textit{first}_k
&= x_k - \tfrac{1}{2}\sigma^2 \bigl({\textstyle \sum_j h_{kj}
\tfrac{\rho'(|x_k - x_j|)}{|x_k - x_j|} (x_k - x_j)
+ \sum_i h_{ik}
\tfrac{\rho'(|x_i - x_k|)}{|x_i - x_k|} (x_i - x_k)}\bigr) \nonumber \\
&= x_k - \sigma^2 \sum_j h_{kj}
\tfrac{\rho'(|x_k - x_j|)}{|x_k - x_j|} (x_k - x_j).
\label{e:tvfirstorder}
\end{align}

The second-order relationship (\ref{e:bridge2}) and calculation
(\ref{e:phihessian}) yield another kernel filter
\begin{equation}
\Hat{\vv{x}}^\textit{second}
= \bigl(\vv{I} - \sigma^2 \mathcal{H}\phi(x)\bigr) \vv{x}
= \bigl(\vv{I} - \tfrac{1}{2} \sigma^2
{\textstyle \sum_i \sum_j h_{ij}
\rho''(|\vv{R}_{ij}\vv{x}|) \vv{R}_{ij}^T \vv{R}_{ij}}
\bigr) \vv{x}
\end{equation}
where the output at pixel $k$ is
\begin{equation}\label{e:tvsecondorder}
\Hat{x}^\textit{second}_k =
x_k - \sigma^2 \sum_j h_{kj} \rho''(|x_k - x_j|) (x_k - x_j).
\end{equation}

We note that in this case, no row-sum normalization is needed since the kernel
filters (\ref{e:tvfirstorder}) and (\ref{e:tvsecondorder}) are already
normalized. Interestingly, the first and second-order kernel filters both have a
form resembling the division-free bilateral filter described in
\cite{milanfar2016new}, and differ only in the factor
$\frac{\rho'(|x_k - x_j|)}{ |x_k - x_j|}$ vs.\ $\rho''(|x_k - x_j|)$.

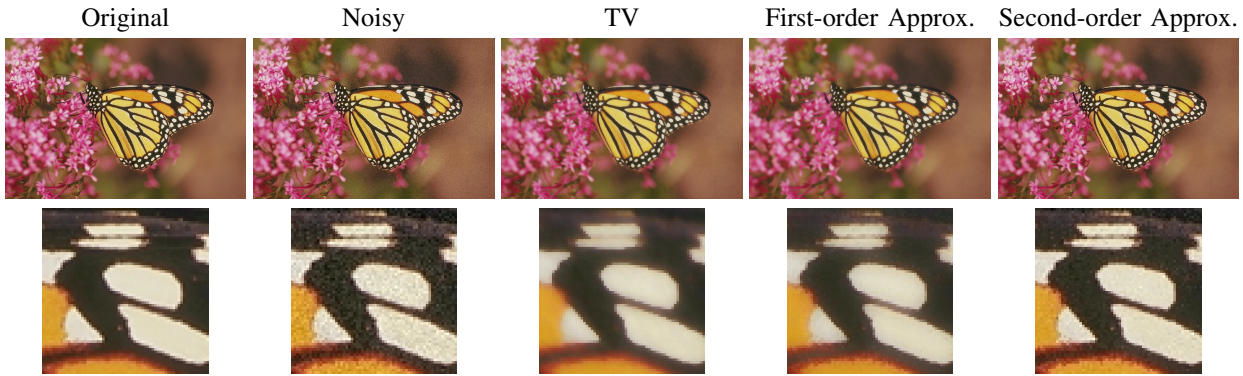
\begin{figure}
\centering
\mbox{\beginpgfgraphicnamed{figures/compare_huber}%
\input{figures/compare_huber.tikz}%
\endpgfgraphicnamed}
\caption{Comparison between Bayesian denoiser with Huber total variation regularization and non-local means with the proposed first and second-order kernel approximation. The first-order kernel approximation achieved PSNR of 37.10 dB, and the second-order kernel approximation achieved PSNR of 29.24 dB.}
\label{fig:compare_huber}
\end{figure}

\begin{figure}
\centering
\includegraphics[width=0.45\linewidth]{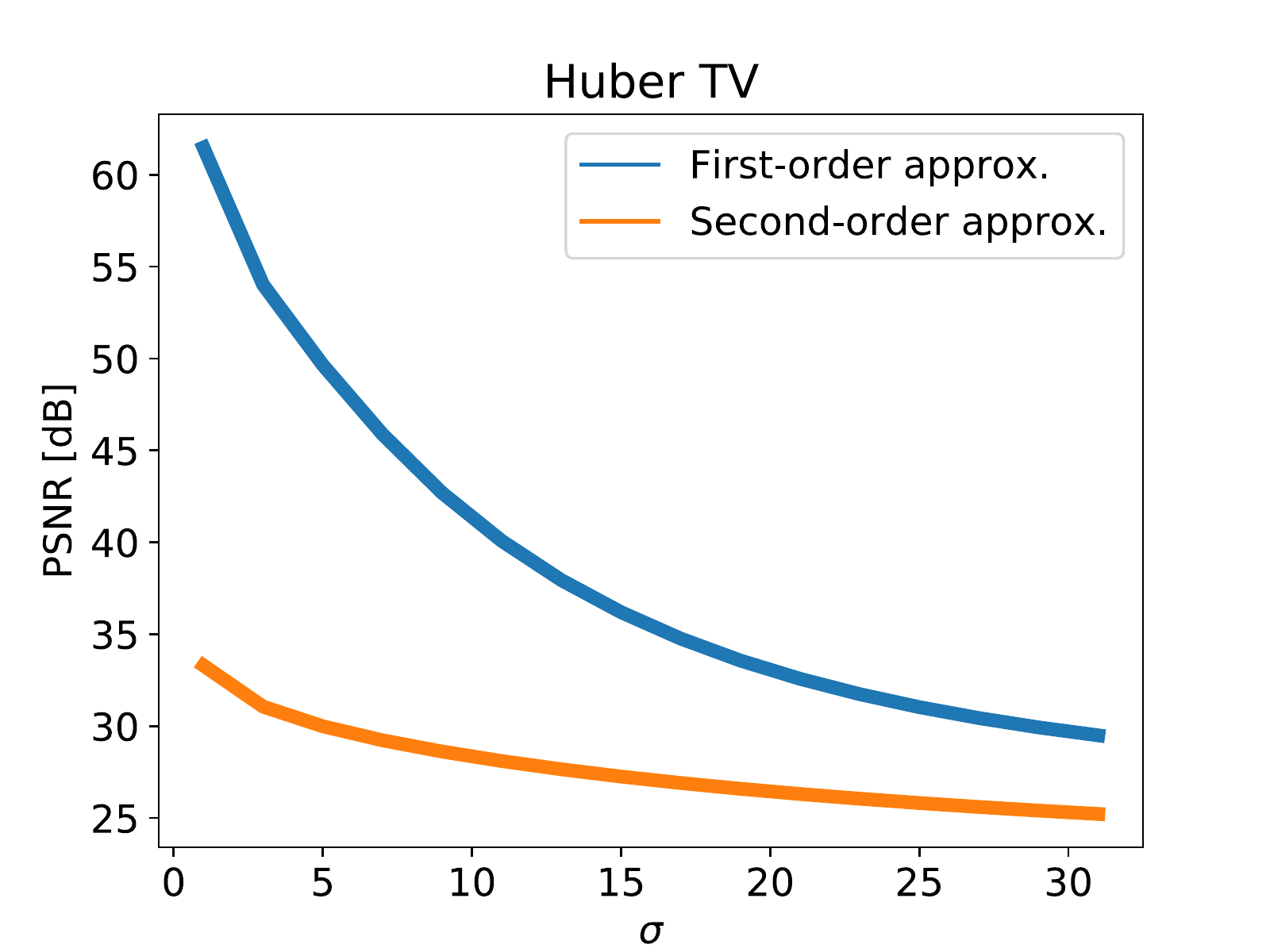}
\caption{PSNR over $\sigma$ for approximating Huber total variation regularized denoising with non-local means.}
\label{fig:huber_psnr}
\end{figure}

In Figure~\ref{fig:compare_huber}, we added white Gaussian noise with standard
deviation of $10$ and applied the first- and second-order kernel filters. The
output with the first-order kernel approximation is both qualitatively and
quantitatively close to the output of total variation denoising, while having
the advantage of being a non-iterative method. Figure~\ref{fig:huber_psnr} shows
the approximation PSNR swept over $\sigma$. The approximation is more accurate
for small $\sigma$.

\subsection{Approximating the Bilateral Filter with First and Second Order Kernel Approximations}
\label{sec:approx}

This section demonstrates the kernel-to-loss direction. Here we begin with a
division-free form of the bilateral filter~\cite{milanfar2016new} (a kernel
filter),
\begin{equation}\label{e:bilateral}
\Hat{x}_i = x_i - \alpha \sum_j h_{ij} k(|x_i - x_j|) (x_i - x_j)
\end{equation}
where $k(t)$ is the range kernel, $h$ is the spatial weight, and $\alpha$ is a
parameter. We approximate it with a MAP problem, using the first- and
second-order relationships to determine the loss function.

We note the similar form of the above bilateral filter to those derived in the
loss-to-kernel direction in the previous experiment with MAP problem
\begin{equation}\label{e:ex2mapproblem}
\operatorname*{arg\,min}_u \,
\frac{1}{2\sigma^2} \|\vv{u} - \vv{x}\|_2^2 + \phi(\vv{u}),
\quad \phi(\vv{u}) = \frac{1}{2} \sum_i \sum_j
h_{ij} \rho(\lvert x_i - x_j \rvert)
\end{equation}
Working backward, (\ref{e:tvfirstorder}) from the first-order relationship
implies
\begin{align}\label{e:lossfirst}
\frac{\rho'(t)}{t} = k(t) \quad
\Rightarrow \quad \rho(t) &= \int \tau k(\tau) \, d\tau + c_0
\end{align}
and (\ref{e:tvsecondorder}) from the second-order relationship implies
\begin{align}\label{e:losssecond}
\rho''(t) = k(t) \qquad
\Rightarrow \quad \rho(t) &= \iint k(\tau) \, d^2\tau  + c_1 t + c_0.
\end{align}
The coefficients $c_0, c_1$ are determined such that $\rho(0) = \rho'(0) = 0$.
To solve (\ref{e:ex2mapproblem}), we use the first-order gradient method, which
guarantees convergence. Since the regularizer for the second order approximation
is smooth and convex, we further add the heavy-ball momentum, which converges to
the optimum at the fastest possible (first-order) convergence
rate~\cite{polyak1964some}. We initialize with the input image.

In the following experiments, we use a Gaussian spatial weighting $h$ that is
$h_{ij} = \exp\bigl(-\|i - j\|^2 / (2\cdot10^2)\bigr)$ over $(i-j) \in
\{-5,\ldots,5\}\times\{-5,\ldots,5\}$ and zero otherwise. The range kernel
$k(t)$ is the boxcar, Gaussian, or exponential kernels. In each case, we sweep
the $\alpha$ parameter and select the value with minimum mean squared error. We
sweep $\alpha$ over $[0.01, 0.2]$ for the first-order approximation and over
$[1, 1024]$ for the second-order approximation. We show the bilateral filter
(\ref{e:bilateral}) and its approximation as a MAP problem
(\ref{e:ex2mapproblem}) where $\rho(t)$ is determined by either the first-order
(\ref{e:lossfirst}) or second-order (\ref{e:losssecond}) relationship.

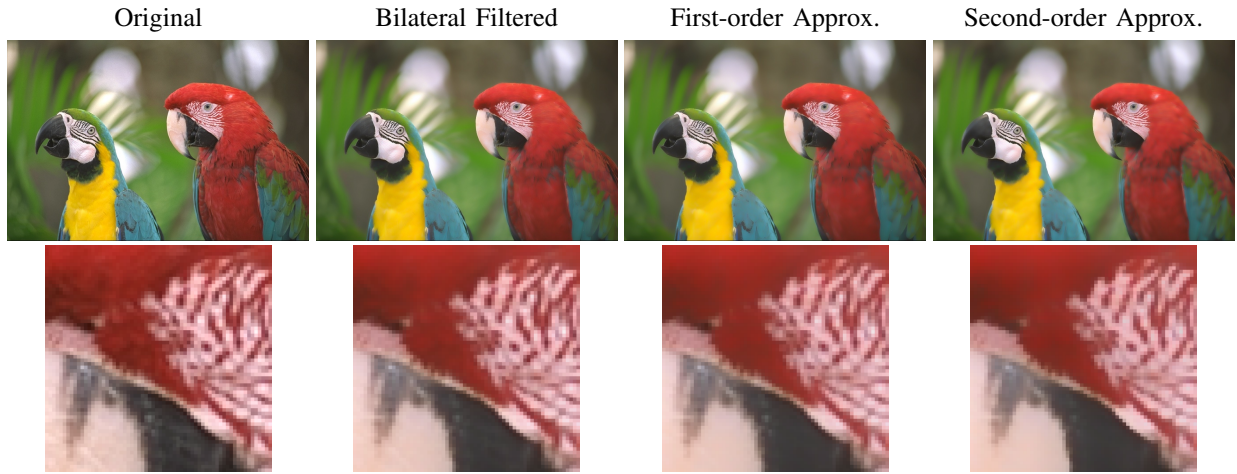
\begin{figure}
\centering
\mbox{\beginpgfgraphicnamed{figures/compare_gaussian}%
\input{figures/compare_gaussian.tikz}%
\endpgfgraphicnamed}
\caption{Comparison between bilateral filter with Gaussian kernel and Bayesian denoiser with the proposed first and second-order kernel approximation. The first-order kernel approximation achieved PSNR of 48.48 dB, and the second-order kernel approximation achieved PSNR of 42.75 dB.}
\label{fig:compare_gaussian}
\end{figure}

\begin{figure}
\centering
\mbox{\beginpgfgraphicnamed{figures/compare_exponential}%
\input{figures/compare_exponential.tikz}%
\endpgfgraphicnamed}
\caption{Comparison between bilateral filter with exponential kernel and Bayesian denoiser with the proposed first and second-order kernel approximation. The first-order kernel approximation achieved PSNR of 46.13 dB, and the second-order kernel approximation achieved PSNR of 44.07 dB.}
\label{fig:compare_exponential}
\end{figure}
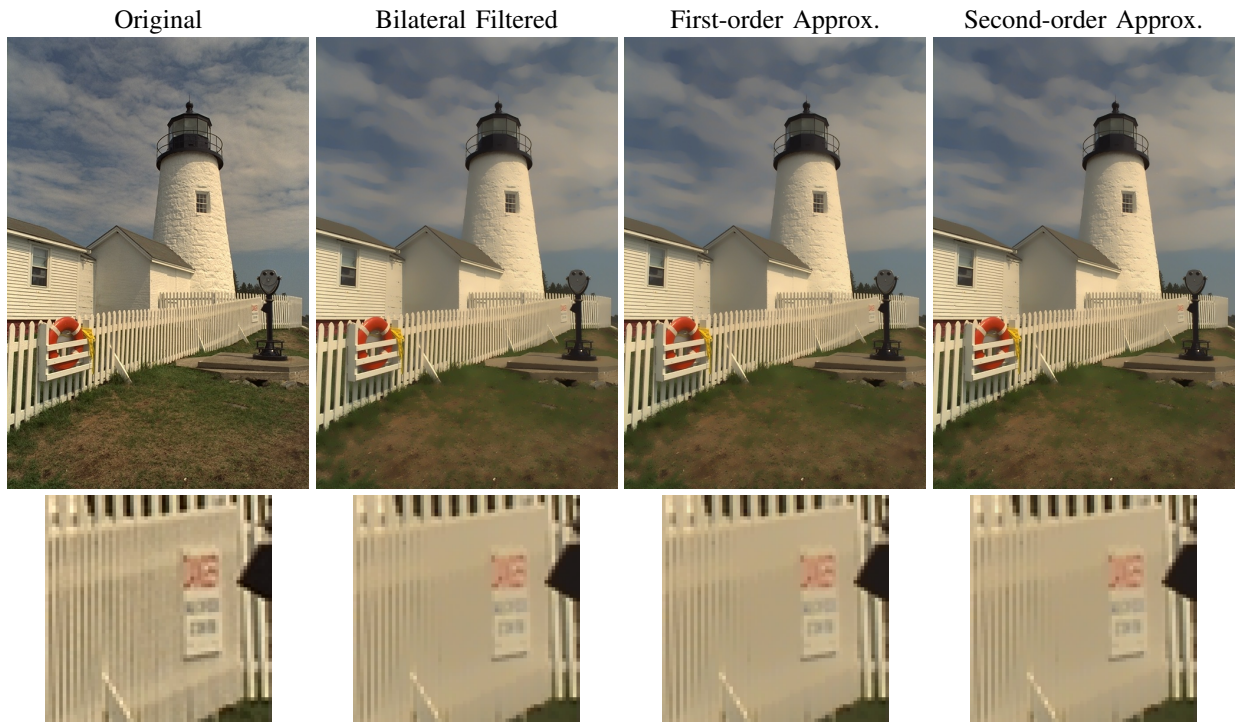

Figures~\ref{fig:compare_gaussian},~\ref{fig:compare_boxcar},
and~\ref{fig:compare_exponential} show the results for $\sigma=30$ along with
the peak signal-to-noise ratio (PSNR) between the MAP approximations to the
bilateral filtered images. We see that both quantitatively and qualitatively,
the images produced from bilateral filtering and derived MAP approximations are
comparable: small details are blurred, while significant features are retained.
Quantitatively, the second-order kernel approximation results in slightly worse
PSNR than the first-order kernel approximation.

\begin{figure}
\centering
\begin{subfigure}[t]{0.45\linewidth}
\includegraphics[width=\linewidth]{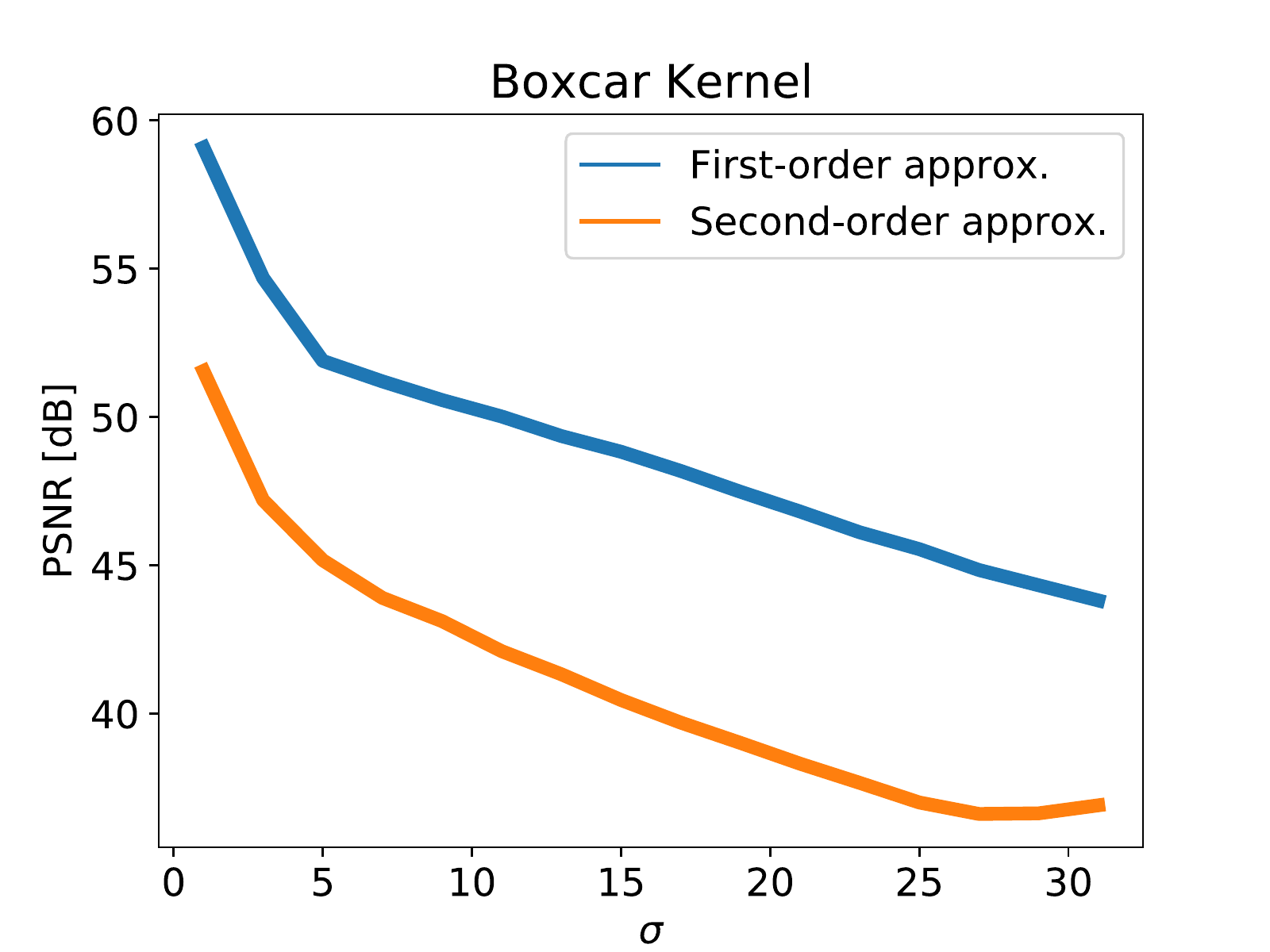}
\end{subfigure}
\begin{subfigure}[t]{0.45\linewidth}
\centering
\includegraphics[width=\linewidth]{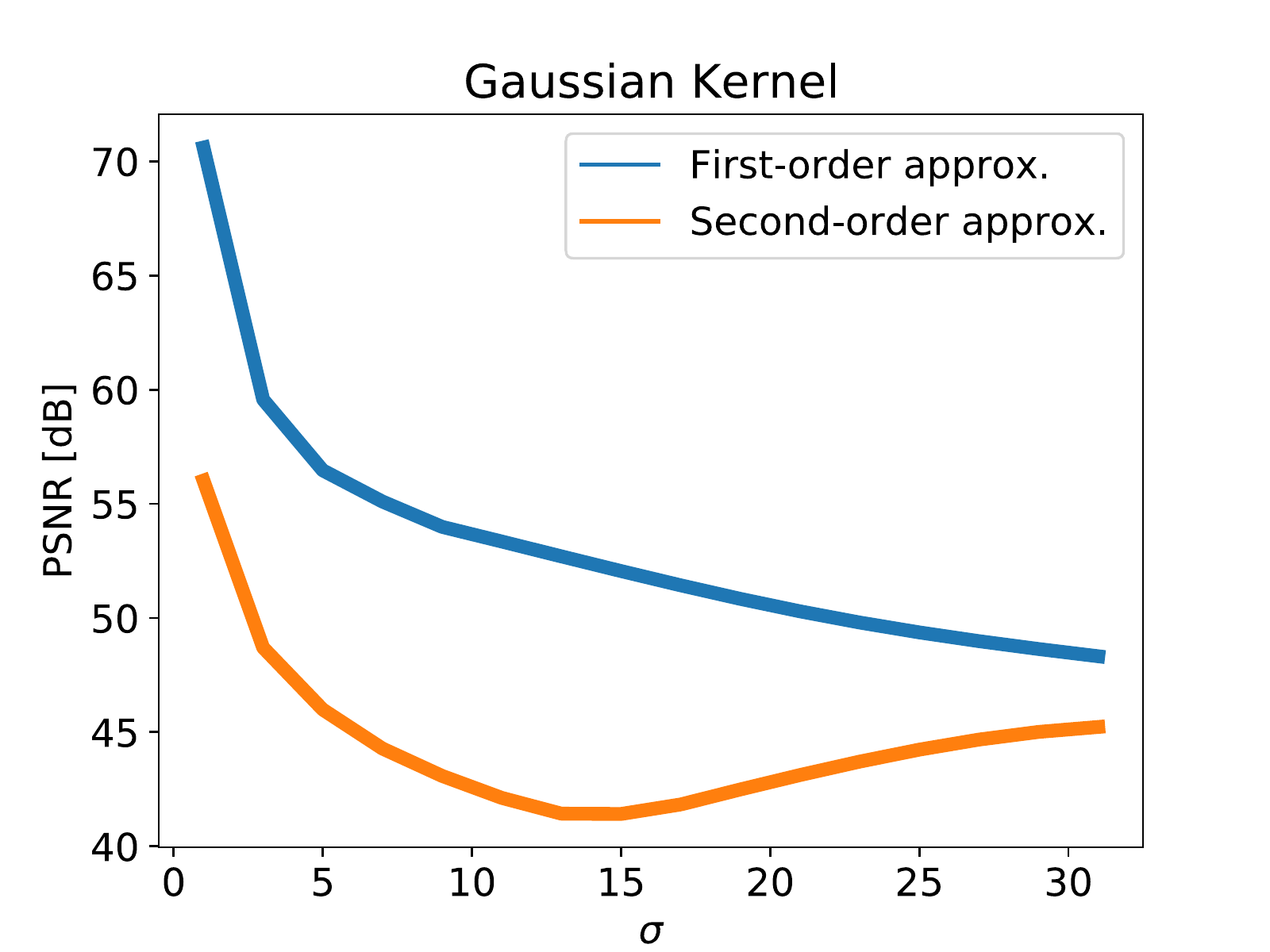}
\end{subfigure}
\newline
\begin{subfigure}[t]{0.45\linewidth}
\centering
\includegraphics[width=\linewidth]{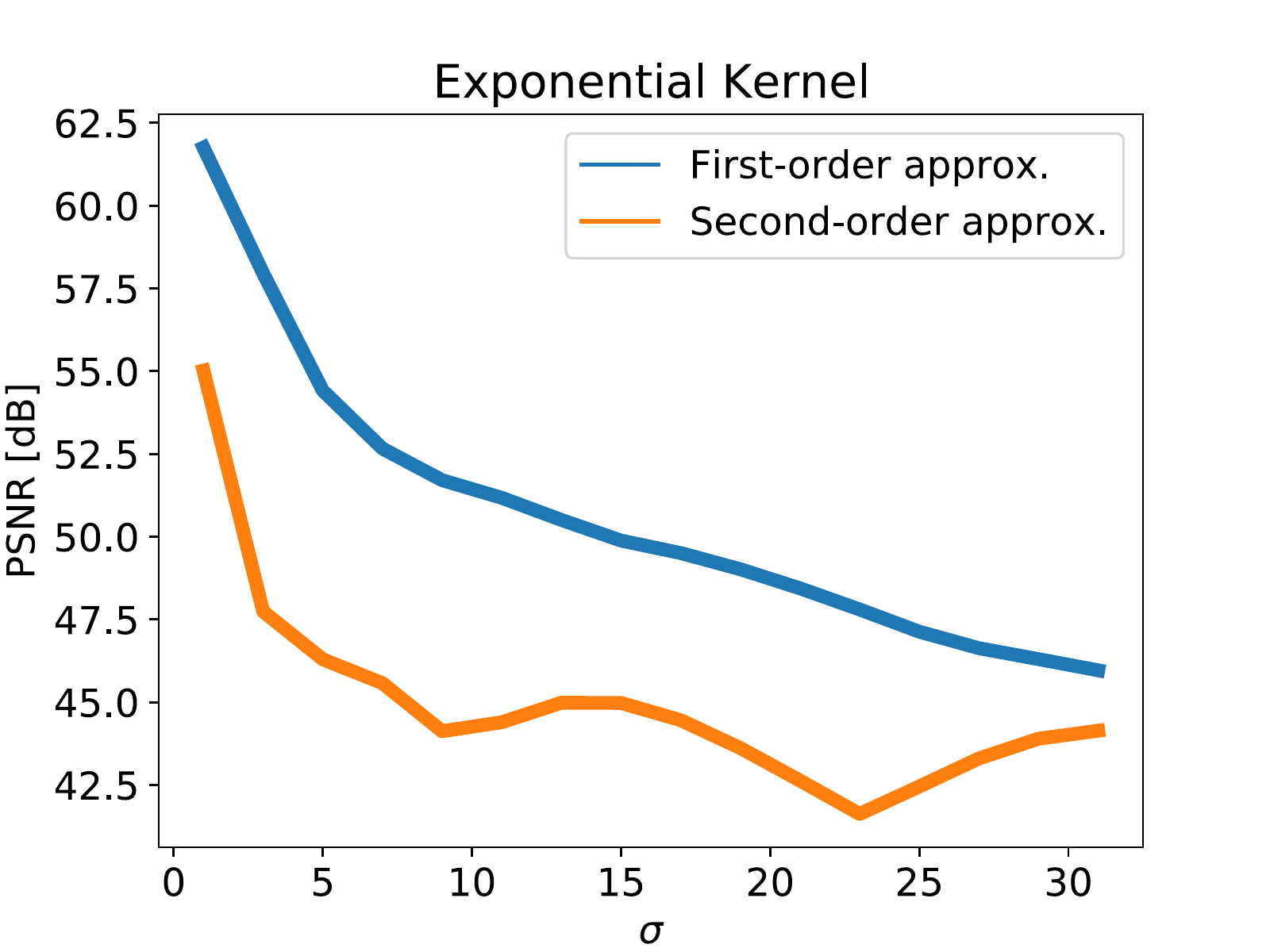}
\end{subfigure}
\caption{PSNR over $\sigma$ for approximating bilateral filter with boxcar,
Gaussian, and exponential kernels with the experimental set up described
in~\ref{sec:approx}.}
\label{fig:kernel_psnr}
\end{figure}

Figure~\ref{fig:kernel_psnr} shows the approximation PSNR swept over $\sigma$
for Gaussian, boxcar, and exponential kernels. As expected, the approximations
are more accurate for small $\sigma$. Also, the first-order kernel approximation
is consistently more accurate than the second-order kernel approximation, even
though the objective function is not convex.

% A peak-signal-to-noise (PSNR) plot between bilateral filter with Gaussian kernel
% and the proposed regularized least squares estimator, while sweeping over the
% regularization parameters $\alpha$ and $\sigma$ is shown in
% Figure~\ref{fig:psnr}. Note that the plots are in log scale in the parameters,
% showing that the regularization is not sensitive to changes in parameters.
% Moreover, even for extremely large parameters, the PSNR achieved is still above
% 39 dB.

% \begin{figure}
% 	\centering
% \begin{subfigure}[t]{0.45\linewidth}
% 	\includegraphics[width=\linewidth]{figures/psnr_sigma}
% \end{subfigure}
% \begin{subfigure}[t]{0.45\linewidth}
% 	\centering
% 	\includegraphics[width=\linewidth]{figures/psnr_alpha}
% \end{subfigure}
% \caption{PSNR plot between bilateral filter with Gaussian kernel and the
% proposed regularized least squares estimator, while sweeping over the
% regularization parameters $\alpha$ and $\sigma$. Note that the plots are in log
% scale in the parameters, showing that the regularization is not sensitive
% to changes in parameters. Moreover, even for extremely large parameters, the
% PSNR achieved is still above 39 dB.}
% \label{fig:psnr}
% \end{figure}

\begin{figure}
\centering
\mbox{\beginpgfgraphicnamed{figures/compare_boxcar}%
\input{figures/compare_boxcar.tikz}%
\endpgfgraphicnamed}
\caption{Comparison between bilateral filter with boxcar kernel and Bayesian denoiser with the proposed first and second-order kernel approximation. The first-order kernel approximation achieved PSNR of 44.09 dB, and the second-order kernel approximation achieved PSNR of 36.79 dB.}
\label{fig:compare_boxcar}
\end{figure}
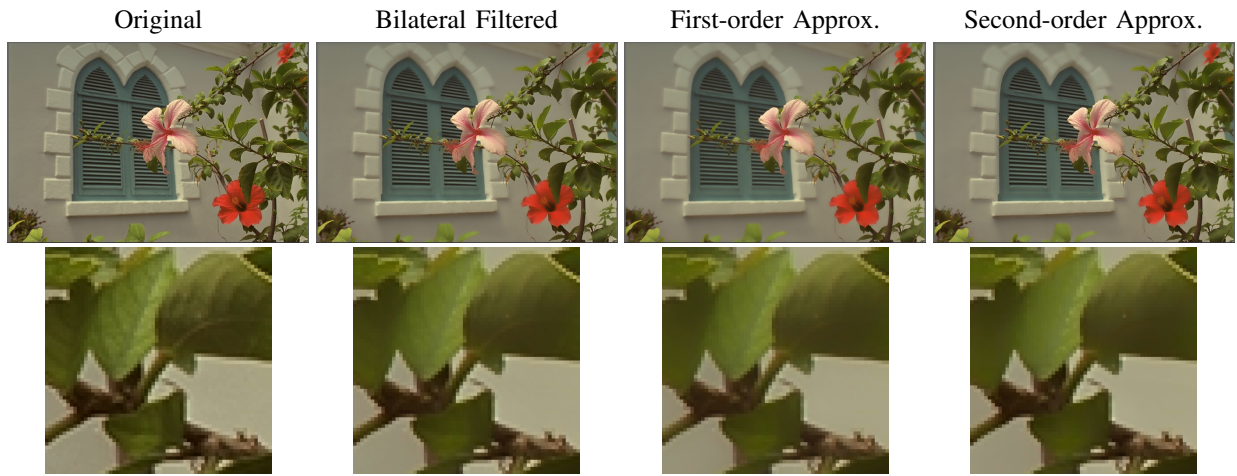

\clearpage
%-------------------------------------------------------------------------------
\section{Final Remarks and Conclusions}

Connections between kernel and regularization-based methods have been studied
before, but for the most part have addressed specific kernels (e.g bilateral) or have not been
directly useful in constructing general, local, non-iterative, and practical filters from
global penalties. In this paper, we established a clear and general way to approximately exchange local kernels $k(\vx)$ and corresponding global regularizers $\phi(\vx)$. In the ``kernelization'' direction ($\phi(\vx)
\rightarrow k(\vx)$) we gave first and second order methods for reducing a global loss to a local kernel filter. The first order loss established a relationship between the tangent space to the regularization function and the (pseudo) linear operator described by the kernel filter; namely
\begin{equation}
\nabla \phi(\vx) = \frac{1}{\sigma^2} \bL(\vx) \vx.
\end{equation}
For the second order approximation, we showed that the Hessian of the regularizer determines the affinity weights $k(\vx)$ as
\begin{equation}
\mathcal{H} \phi(\vx) = \frac{1}{\sigma^2} \bL(\vx) \approx
\frac{\alpha}{\sigma^2} \left(\mD(\vx) - \mK(\vx)\right).
\end{equation}
In the other direction ($k(\vx) \rightarrow \phi(\vx)$) we enabled a Bayesian interpretation or justification for the use of a positive-definite kernel for locally-adapted filtering, and provided a practical path to computing the implied penalties $\phi(\vx)$.

Variational/Bayesian/proximal formulations often result in optimization problems
that do not have closed-form solutions, and therefore typically require {\em
global} iterative solutions. Our contribution here is to establish how one can
approximate the solution of the resulting global optimization problems with use
of ``kernelized'' locally adaptive filters which approximates the global
solution in one-shot, using only local operations.

This is significant and timely, particularly with respect to the wide use of
proximal operators in learning, optimization theory, and their use in solving
inverse problems. In light of our work, approaches that replace proximal
operator in ad hoc fashion with simpler (often patch-based) denoisers can be
replaced with this more principled approach that makes the calculation of the
appropriate kernel filters more direct and transparent.

\clearpage
\appendix

\section{Calculation of filters for Dirichlet energy loss }\label{sec:wn_for_h1}
Point-wise we have the gradient of the approximate filter as:
\begin{equation} \label{eq:localconvfilter}
\begin{aligned}
\widehat{x}^\mathit{approx}_n
&= x_n - \sigma^2 \nabla \phi(\vv{x})_n \\
&= 2\sigma^2 x_{n+1} + (1 - 4\sigma^2) x_n + 2\sigma^2 x_{n-1}.
\end{aligned}
\end{equation}
The solution of (\ref{e:h1_minimization}) is approximated by cyclic convolution
with the 3-tap filter $\left[\sigma^2, \: (1 - 2\sigma^2), \: \sigma^2 \right]$,
which for sufficiently small\footnote{Specifically, for $\sigma\leq
\frac{1}{2\sqrt{2}}$, which is a zero of the 3-tap filter defined by the
approximation.} $\sigma$, is strictly a smoothing filter as intended.

Now let us compute the exact MAP solution by passing to the Fourier domain. Specifically, under the DFT, the objective decouples over spectral coefficients,
\begin{equation}
\Hat{X}_k =
\operatorname*{argmin}_{U_k}
\frac{1}{2\sigma^2}|U_k - X_k|^2 + |\mathrm{e}^{i2\pi k/N} - 1|^2 |U_k|^2,
\end{equation}
where $U_k$ and $X_k$ denote the $k$th spectral coefficients of $\vv{u}$ and
$\vv{x}$. The $|\mathrm{e}^{i2\pi k/N} - 1|^2$ factor simplifies to
$2(1-\cos\tfrac{2\pi k}{N})$. Differentiating and solving the Euler--Lagrange
equation, the MAP solution $\vv{\Hat{X}}$ is the pointwise product (in the frequency domain) of $\vv{X}$ with a denoising filter spectral coefficients,
\begin{equation}
\Hat{X}_k = W_k X_k, \quad
W_k = \bigl(1 + 4\sigma^2 (1 - \cos \tfrac{2\pi k}{N})\bigr)^{-1}.
\end{equation}
Back in the spatial domain, the exact MAP solution is the cyclic convolution
\begin{equation}
\Hat{x}_n = \sum_{m=0}^{N-1} w_m x_{n-m}.
\end{equation}
where $w_n$ is the inverse transform of $W_k$, given by
\begin{equation}
w_n = \frac{1}{N} \sum_{k=0}^{N-1} W_k \mathrm{e}^{i2\pi k n/N}
= \frac{1 - r}{(1 + r)(1 - r^N)} (r^{n} + r^{N-n})
\end{equation}
with $r = 1 + (1 - \sqrt{1 + 8 \sigma^2})/(4\sigma^2)$. The details of the calculation are as follows:

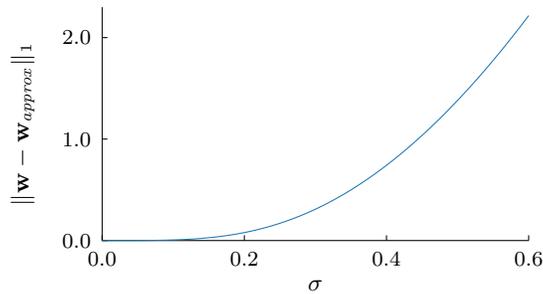
\begin{figure}[!t]
\centering%
\mbox{%
\beginpgfgraphicnamed{figures/h1_distance}%
\input{figures/h1_distance.tikz}%
\endpgfgraphicnamed}
\caption{\label{fig:h1_distance}$\ell^1$ distance between exact and approximate
$H^1$ denoising filters vs.\ $\sigma$.}
\end{figure}

Under the DFT,
\begin{align*}
W_k &= \frac{1}{1 + 4\sigma^2 (1 - \cos \tfrac{2\pi k}{N})} \\
% &= \frac{1}{1 + 4\sigma^2 - 2\sigma^2 \mathrm{e}^{i 2\pi k/N}
% - 2\sigma^2 \mathrm{e}^{-i 2\pi k/N}} \\
&= \frac{-\mathrm{e}^{-i 2\pi k/N}}
{2\sigma^2 (1 - r \mathrm{e}^{-i 2\pi k/N})
(1 - r^{-1} \mathrm{e}^{-i 2\pi k/N})} \\
&= \frac{\mathrm{e}^{-i 2\pi k/N}}{2\sigma^2 (1 - r^2)}
\left(\frac{r^2}{1 - r \mathrm{e}^{-i 2\pi k/N}}
- \frac{1}{1 - r^{-1} \mathrm{e}^{-i 2\pi k/N}}\right)
\end{align*}
where $r = 1 + (1 - \sqrt{1 + 8 \sigma^2})/(4\sigma^2)$. Therefore by the
transform pair $\operatorname{DFT}\bigl(r^n/(1 - r^N)\bigr) = (1 - r
\mathrm{e}^{-i 2\pi k/N})^{-1}$, the inverse transform of $W_k$ is
\begin{align*}
w_n &=
\frac{1}{2\sigma^2 (1 - r^2)}\left(
\frac{r^2}{1 - r^N} r^{n - 1} - \frac{1}{1 - r^{-N}} r^{-n+1} \right) \\
% &= \frac{r}{2\sigma^2 (1 - r^2)(1 - r^N)}\left(
% r^{n} + r^{N-n} \right) \\
&= \frac{1 - r}{(1 + r)(1 - r^N)}\left(
r^{n} + r^{N-n} \right).
\end{align*}

\bibliographystyle{IEEEtran}
\bibliography{paper}

\end{document}

%% file: figures/h1_filters.tikz
% TikZ Figure: Comparison of H1 denoising filters.

\begin{tikzpicture}[scale=1.45,xscale=0.25]
\pgfdeclareplotmark{o}
{
  \pgftransformresetnontranslations
  \pgftransformscale{0.038} % This value controls marker size
  \fill (0,0) circle (1.16);
}
\pgfdeclareplotmark{x}
{
  \pgftransformresetnontranslations
  \pgftransformscale{0.038} % This value controls marker size
  \draw [line width=0.85pt,line cap=round]
  (-1,1) -- (1,-1) (-1,-1) -- (1,1);
}

\begin{scope}
\draw (-5,1.2) -- (-5,0) -- (5,0) node [midway,below=11pt] {\footnotesize $n$};
\foreach \x in {-5, 0, 5}
{
  \begin{scope}[xshift=\x cm]
    \pgftransformresetnontranslations
    \draw (0,2pt) -- (0,0) node [below] {\scriptsize $\x$};
  \end{scope}
}
\foreach \y in {0.0, 0.5, 1.0}
{
  \begin{scope}[xshift=-5cm,yshift=\y cm]
    \pgftransformresetnontranslations
    \draw (2pt,0) -- (0,0) node [left] {\scriptsize $\y$};
  \end{scope}
}
% sigma = 0.2 (L1 dist = 0.0787399).
\draw (0, 1.2) node [above] {
\small $\sigma = 0.2$};
\draw [PlotColorA] plot [mark=o] coordinates {(-5,0) (-4,0) (-3,0.0003) (-2,0.0042) (-1,0.0603) (0,0.8704) (1,0.0603) (2,0.0042) (3,0.0003) (4,0) (5,0)};
\draw [PlotColorD] plot [mark=x] coordinates {(-5,0) (-4,0) (-3,0) (-2,0) (-1,0.0800) (0,0.8400) (1,0.0800) (2,0) (3,0) (4,0) (5,0)};
\end{scope}

\begin{scope}[xshift=14cm]
\draw (-5,1.2) -- (-5,0) -- (5,0) node [midway,below=11pt] {\footnotesize $n$};
\foreach \x in {-5, 0, 5}
{
  \begin{scope}[xshift=\x cm]
    \pgftransformresetnontranslations
    \draw (0,2pt) -- (0,0) node [below] {\scriptsize $\x$};
  \end{scope}
}
\foreach \y in {0.0, 0.5, 1.0}
{
  \begin{scope}[xshift=-5cm,yshift=\y cm]
    \pgftransformresetnontranslations
    \draw (2pt,0) -- (0,0) node [left] {\scriptsize $\y$};
  \end{scope}
}
% sigma = 0.3 (L1 dist = 0.308997).
\draw (0, 1.2) node [above] {
\small $\sigma = 0.3$};
\draw [PlotColorA] plot [mark=o] coordinates {(-5,0) (-4,0.0003) (-3,0.0019) (-2,0.0138) (-1,0.1028) (0,0.7625) (1,0.1028) (2,0.0138) (3,0.0019) (4,0.0003) (5,0)};
\draw [PlotColorD] plot [mark=x] coordinates {(-5,0) (-4,0) (-3,0) (-2,0) (-1,0.1800) (0,0.6400) (1,0.1800) (2,0) (3,0) (4,0) (5,0)};

% Legend
\begin{scope}[xshift=1.75cm,yshift=0.95cm,xscale=0.55,yscale=0.25]
\draw [PlotColorA,yshift=0cm]
plot [mark=o,mark repeat=2,mark phase=2] coordinates
{(0,0) (1,0) (2,0)} node [right,black]
{\scriptsize $\vv{w}$};

\draw [PlotColorD,yshift=-1cm]
plot [mark=x,mark repeat=2,mark phase=2] coordinates
{(0,0) (1,0) (2,0)} node [right,black]
{\scriptsize $\vv{w}_\mathit{approx}$};
\end{scope}
\end{scope}

\begin{scope}[yshift=-2.4cm]
\draw (-5,1.2) -- (-5,0) -- (5,0) node [midway,below=11pt] {\footnotesize $n$};
\foreach \x in {-5, 0, 5}
{
  \begin{scope}[xshift=\x cm]
    \pgftransformresetnontranslations
    \draw (0,2pt) -- (0,0) node [below] {\scriptsize $\x$};
  \end{scope}
}
\foreach \y in {0.0, 0.5, 1.0}
{
  \begin{scope}[xshift=-5cm,yshift=\y cm]
    \pgftransformresetnontranslations
    \draw (2pt,0) -- (0,0) node [left] {\scriptsize $\y$};
  \end{scope}
}
% sigma = 0.4 (L1 dist = 0.741772).
\draw (0, 1.2) node [above] {
\small $\sigma = 0.4$};
\draw [PlotColorA] plot [mark=o] coordinates {(-5,0.0002) (-4,0.0011) (-3,0.0056) (-2,0.0273) (-1,0.1346) (0,0.6623) (1,0.1346) (2,0.0273) (3,0.0056) (4,0.0011) (5,0.0002)};
\draw [PlotColorD] plot [mark=x] coordinates {(-5,0) (-4,0) (-3,0) (-2,0) (-1,0.3200) (0,0.3600) (1,0.3200) (2,0) (3,0) (4,0) (5,0)};
\end{scope}

\begin{scope}[xshift=14cm,yshift=-2.4cm]
\draw (-5,1.2) -- (-5,0) -- (5,0) node [midway,below=11pt] {\footnotesize $n$};
\foreach \x in {-5, 0, 5}
{
  \begin{scope}[xshift=\x cm]
    \pgftransformresetnontranslations
    \draw (0,2pt) -- (0,0) node [below] {\scriptsize $\x$};
  \end{scope}
}
\foreach \y in {0.0, 0.5, 1.0}
{
  \begin{scope}[xshift=-5cm,yshift=\y cm]
    \pgftransformresetnontranslations
    \draw (2pt,0) -- (0,0) node [left] {\scriptsize $\y$};
  \end{scope}
}
% sigma = 0.45 (L1 dist = 1.03619).
\draw (0, 1.2) node [above] {
\small $\sigma = 0.45$};
\draw [PlotColorA] plot [mark=o] coordinates {(-5,0.0005) (-4,0.0019) (-3,0.0081) (-2,0.0345) (-1,0.1460) (0,0.6178) (1,0.1460) (2,0.0345) (3,0.0081) (4,0.0019) (5,0.0005)};
\draw [PlotColorD] plot [mark=x] coordinates {(-5,0) (-4,0) (-3,0) (-2,0) (-1,0.4050) (0,0.1900) (1,0.4050) (2,0) (3,0) (4,0) (5,0)};
\end{scope}
\end{tikzpicture}

%% file: figures/kernels.tikz
% TikZ Figure: Examples of kernel functions.

\begin{tikzpicture}[scale=1,yscale=2.4,yscale=1.1]
\draw (0,1) node [above=3pt] {\small Kernels $k(t)$};

\draw (-3,1) -- (-3,0) -- (3,0) node [midway,below=11pt] {\footnotesize $t$};
\foreach \x in {-3, -2, -1, 0, 1, 2, 3}
{
  \begin{scope}[xshift=\x cm]
    \pgftransformresetnontranslations
    \draw (0,2pt) -- (0,0) node [below] {\scriptsize $\x$};
  \end{scope}
}
\foreach \y in {0.0, 0.5, 1.0}
{
  \begin{scope}[xshift=-3cm,yshift=\y cm]
    \pgftransformresetnontranslations
    \draw (2pt,0) -- (0,0) node [left] {\scriptsize $\y$};
  \end{scope}
}

% Boxcar
\draw [PlotColorA,thick] plot coordinates {(-3,0) (-1,0) (-1,1) (1,1) (1,0)
(3,0)};
% Gaussian
\draw [PlotColorB,thick] plot [smooth] coordinates {(-3,0.011)
(-2.667,0.029) (-2.5,0.044) (-2.333,0.066) (-2.083,0.114) (-1.917,0.159)
(-1.833,0.186) (-1.667,0.249) (-1.417,0.367) (-1.167,0.506) (-0.75,0.755)
(-0.583,0.844) (-0.5,0.882) (-0.333,0.946) (-0.25,0.969) (-0.167,0.986)
(-0.083,0.997) (0,1) (0.083,0.997) (0.167,0.986) (0.25,0.969) (0.333,0.946)
(0.5,0.882) (0.583,0.844) (0.75,0.755) (1.167,0.506) (1.417,0.367) (1.667,0.249)
(1.833,0.186) (1.917,0.159) (2.083,0.114) (2.333,0.066) (2.5,0.044)
(2.667,0.029) (3,0.011)};
% Cauchy
\draw [PlotColorC,thick] plot [smooth] coordinates {(-3,0.182) (-2.667,0.22)
(-2.333,0.269) (-2,0.333) (-1.75,0.395) (-1.5,0.471) (-1.25,0.561) (-1,0.667)
(-0.583,0.855) (-0.417,0.92) (-0.25,0.97) (-0.167,0.986) (-0.083,0.997) (0,1)
(0.083,0.997) (0.167,0.986) (0.25,0.97) (0.417,0.92) (0.583,0.855) (1,0.667)
(1.25,0.561) (1.5,0.471) (1.75,0.395) (2,0.333) (2.333,0.269) (2.667,0.22)
(3,0.182)};
% Exponential
\draw [PlotColorD,thick] plot [smooth] coordinates {(-3,0.12) (-2.5,0.171)
(-2.25,0.204) (-2,0.243) (-1.75,0.29) (-1.583,0.326) (-1.25,0.413)
(-1.083,0.465) (-0.917,0.523) (-0.75,0.588) (-0.583,0.662) (-0.417,0.745)
(-0.25,0.838) (0,1)}
plot [smooth] coordinates {(0,1) (0.25,0.838) (0.417,0.745) (0.583,0.662)
(0.75,0.588) (0.917,0.523) (1.083,0.465) (1.25,0.413) (1.583,0.326) (1.75,0.29)
(2,0.243) (2.25,0.204) (2.5,0.171) (3,0.12)};

% Legend
\begin{scope}[xshift=1.85cm,yshift=0.95cm,xscale=0.35,yscale=0.12,thick]
\draw [PlotColorA,yshift=0cm] plot coordinates
{(0,0) (1,0)} node [right,black] {\scriptsize Boxcar\vphantom{y}};
\draw [PlotColorB,yshift=-1cm] plot coordinates
{(0,0) (1,0)} node [right,black] {\scriptsize Gaussian\vphantom{y}};
\draw [PlotColorC,yshift=-2cm] plot coordinates
{(0,0) (1,0)} node [right,black] {\scriptsize Cauchy\vphantom{y}};
\draw [PlotColorD,yshift=-3cm] plot coordinates
{(0,0) (1,0)} node [right,black] {\scriptsize Exponential\vphantom{y}};
\end{scope}
\end{tikzpicture}

%% file: figures/losses.tikz
% TikZ Figure: Examples of loss functions rho.

\begin{tikzpicture}[scale=1,yscale=2.4,yscale=1.1]

\draw (0,1) node [above=3pt] {\small Losses $\rho(t)$};

\draw (-3,1) -- (-3,0) -- (3,0) node [midway,below=11pt] {\footnotesize $t$};
\foreach \x in {-3, -2, -1, 0, 1, 2, 3}
{
  \begin{scope}[xshift=\x cm]
    \pgftransformresetnontranslations
    \draw (0,2pt) -- (0,0) node [below] {\scriptsize $\x$};
  \end{scope}
}
\foreach \y in {0.0, 0.5, 1.0}
{
  \begin{scope}[xshift=-3cm,yshift=\y cm]
    \pgftransformresetnontranslations
    \draw (2pt,0) -- (0,0) node [left] {\scriptsize $\y$};
  \end{scope}
}

% Boxcar
\draw [PlotColorA,thick] plot coordinates {(-3,0.5) (-1,0.5)}
plot [smooth] coordinates {(-1,0.5) (-0.833,0.347) (-0.75,0.281) (-0.583,0.17)
(-0.5,0.125) (-0.417,0.087) (-0.333,0.056) (-0.25,0.031) (-0.167,0.014)
(-0.083,0.003) (0,0) (0.083,0.003) (0.167,0.014) (0.25,0.031) (0.333,0.056)
(0.417,0.087) (0.5,0.125) (0.583,0.17) (0.75,0.281) (0.917,0.42) (1,0.5)}
plot coordinates {(1,0.5) (3,0.5)};
% Gaussian
\draw [PlotColorB,thick] plot [smooth] coordinates {(-3,0.494)
(-2.667,0.486) (-2.333,0.467) (-2.083,0.443) (-1.833,0.407) (-1.667,0.375)
(-1.417,0.317) (-1.167,0.247) (-0.75,0.123) (-0.5,0.059) (-0.333,0.027)
(-0.25,0.015) (-0.083,0.002) (0,0) (0.083,0.002) (0.25,0.015) (0.333,0.027)
(0.5,0.059) (0.75,0.123) (1.167,0.247) (1.417,0.317) (1.667,0.375) (1.833,0.407)
(2.083,0.443) (2.333,0.467) (2.667,0.486) (3,0.494)};
% Cauchy
\draw [PlotColorC,thick] plot [smooth] coordinates {(-3,0.852)
(-2.583,0.734) (-2.167,0.604) (-1.75,0.464) (-1.083,0.231) (-0.75,0.124)
(-0.5,0.059) (-0.333,0.027) (-0.167,0.007) (0,0) (0.167,0.007) (0.333,0.027)
(0.5,0.059) (0.75,0.124) (1.083,0.231) (1.75,0.464) (2.167,0.604) (2.583,0.734)
(3,0.852)};
% Exponential
\draw [PlotColorD,thick] plot [smooth] coordinates {(-3,0.622) (-2.5,0.586)
(-2,0.535) (-1.583,0.476) (-1.25,0.415) (-0.917,0.337) (-0.75,0.291)
(-0.583,0.239) (-0.417,0.18) (-0.25,0.115) (0,0)}
plot [smooth] coordinates {(0,0) (0.25,0.115) (0.417,0.18) (0.583,0.239)
(0.75,0.291) (0.917,0.337) (1.25,0.415) (1.583,0.476) (2,0.535) (2.5,0.586)
(3,0.622)};
\end{tikzpicture}

%% file: figures/kernels_barron.tikz
% TikZ Figure: Examples of Barron's kernel functions.

\begin{tikzpicture}[scale=1,yscale=2.4]
\draw (0,1.1) node [above=3pt] {\small Kernels $k(t)$};

\draw (-3,1.1) -- (-3,0) -- (3,0) node [midway,below=11pt] {\footnotesize $t$};
\foreach \x in {-3, -2, -1, 0, 1, 2, 3}
{
  \begin{scope}[xshift=\x cm]
    \pgftransformresetnontranslations
    \draw (0,2pt) -- (0,0) node [below] {\scriptsize $\x$};
  \end{scope}
}
\foreach \y in {0.0, 0.5, 1.0}
{
  \begin{scope}[xshift=-3cm,yshift=\y cm]
    \pgftransformresetnontranslations
    \draw (2pt,0) -- (0,0) node [left] {\scriptsize $\y$};
  \end{scope}
}

% beta = -inf
\draw [PlotColorA,thick] plot [smooth] coordinates {(-3,0.011) (-2.667,0.029)
(-2.5,0.044) (-2.333,0.066) (-2.083,0.114) (-1.917,0.159) (-1.833,0.186)
(-1.667,0.249) (-1.417,0.367) (-1.167,0.506) (-0.75,0.755) (-0.583,0.844)
(-0.5,0.882) (-0.333,0.946) (-0.25,0.969) (-0.167,0.986) (-0.083,0.997) (0,1)
(0.083,0.997) (0.167,0.986) (0.25,0.969) (0.333,0.946) (0.5,0.882) (0.583,0.844)
(0.75,0.755) (1.167,0.506) (1.417,0.367) (1.667,0.249) (1.833,0.186)
(1.917,0.159) (2.083,0.114) (2.333,0.066) (2.5,0.044) (2.667,0.029) (3,0.011)};
% beta = -2
\draw [PlotColorB,thick] plot [smooth] coordinates {(-3,0.095) (-2.667,0.13)
(-2.333,0.179) (-2.083,0.23) (-1.833,0.295) (-1.583,0.378) (-1.333,0.479)
(-1.083,0.598) (-0.667,0.81) (-0.5,0.886) (-0.417,0.919) (-0.25,0.969)
(-0.167,0.986) (-0.083,0.997) (0,1) (0.083,0.997) (0.167,0.986) (0.25,0.969)
(0.417,0.919) (0.5,0.886) (0.667,0.81) (1.083,0.598) (1.333,0.479) (1.583,0.378)
(1.833,0.295) (2.083,0.23) (2.333,0.179) (2.667,0.13) (3,0.095)};
% beta = -1
\draw [PlotColorC,thick] plot [smooth] coordinates {(-3,0.125) (-2.667,0.162)
(-2.333,0.212) (-2.167,0.243) (-2,0.281) (-1.75,0.348) (-1.5,0.432)
(-1.25,0.533) (-1.083,0.609) (-0.667,0.813) (-0.5,0.887) (-0.417,0.919)
(-0.25,0.97) (-0.167,0.986) (-0.083,0.997) (0,1) (0.083,0.997) (0.167,0.986)
(0.25,0.97) (0.417,0.919) (0.5,0.887) (0.667,0.813) (1.083,0.609) (1.25,0.533)
(1.5,0.432) (1.75,0.348) (2,0.281) (2.167,0.243) (2.333,0.212) (2.667,0.162)
(3,0.125)};
% beta = 0
\draw [PlotColorD,thick] plot [smooth] coordinates {(-3,0.033) (-2.5,0.059)
(-2.25,0.08) (-2.083,0.1) (-1.917,0.124) (-1.75,0.156) (-1.583,0.197)
(-1.5,0.221) (-1.333,0.28) (-1.25,0.315) (-1.083,0.397) (-1,0.444)
(-0.833,0.551) (-0.5,0.79) (-0.333,0.898) (-0.25,0.94) (-0.167,0.973)
(-0.083,0.993) (0,1) (0.083,0.993) (0.167,0.973) (0.25,0.94) (0.333,0.898)
(0.5,0.79) (0.833,0.551) (1,0.444) (1.083,0.397) (1.25,0.315) (1.333,0.28)
(1.5,0.221) (1.583,0.197) (1.75,0.156) (1.917,0.124) (2.083,0.1) (2.25,0.08)
(2.5,0.059) (3,0.033)};
% beta = 0.5
\draw [PlotColorE,thick] plot [smooth] coordinates {(-3,0.232) (-2.667,0.27)
(-2.333,0.317) (-2,0.377) (-1.75,0.434) (-1.5,0.503) (-1.25,0.585) (-1,0.682)
(-0.583,0.858) (-0.417,0.921) (-0.25,0.97) (-0.167,0.986) (-0.083,0.997) (0,1)
(0.083,0.997) (0.167,0.986) (0.25,0.97) (0.417,0.921) (0.583,0.858) (1,0.682)
(1.25,0.585) (1.5,0.503) (1.75,0.434) (2,0.377) (2.333,0.317) (2.667,0.27)
(3,0.232)};
% beta = 1
\draw [PlotColorF,thick] plot [smooth] coordinates {(-3,0.316) (-2.583,0.361)
(-2.25,0.406) (-1.917,0.463) (-1.667,0.514) (-1.417,0.577) (-1.167,0.651)
(-0.917,0.737) (-0.5,0.894) (-0.333,0.949) (-0.167,0.986) (-0.083,0.997) (0,1)
(0.083,0.997) (0.167,0.986) (0.333,0.949) (0.5,0.894) (0.917,0.737)
(1.167,0.651) (1.417,0.577) (1.667,0.514) (1.917,0.463) (2.25,0.406)
(2.583,0.361) (3,0.316)};
% beta = 2
\draw [PlotColorG,thick] plot coordinates {(-3,1) (3,1)};

% Legend
\begin{scope}[xshift=-0.42cm,yshift=0.7cm,xscale=0.32,yscale=0.1,thick]
\draw [PlotColorA,yshift=0cm] plot coordinates
{(0,0) (1,0)} node [right,black] {\tiny $-\infty$};
\draw [PlotColorB,yshift=-1cm] plot coordinates
{(0,0) (1,0)} node [right,black] {\tiny $-2$};
\draw [PlotColorC,yshift=-2cm] plot coordinates
{(0,0) (1,0)} node [right,black] {\tiny $-1$};
\draw [PlotColorD,yshift=-3cm] plot coordinates
{(0,0) (1,0)} node [right,black] {\tiny $0$};
\draw [PlotColorE,yshift=-4cm] plot coordinates
{(0,0) (1,0)} node [right,black] {\tiny $1/2$};
\draw [PlotColorF,yshift=-5cm] plot coordinates
{(0,0) (1,0)} node [right,black] {\tiny $1$};
\draw [PlotColorG,yshift=-6cm] plot coordinates
{(0,0) (1,0)} node [right,black] {\tiny $2$};
\end{scope}
\end{tikzpicture}

%% file: figures/losses_barron.tikz
% TikZ Figure: Examples of loss functions rho corresponding to Barron's kernels.

\begin{tikzpicture}[scale=1,yscale=1.2,yscale=1.1]

\draw (0,2) node [above=3pt] {\small Losses $\rho(t)$};

\draw (-3,2) -- (-3,0) -- (3,0) node [midway,below=11pt] {\footnotesize $t$};
\foreach \x in {-3, -2, -1, 0, 1, 2, 3}
{
  \begin{scope}[xshift=\x cm]
    \pgftransformresetnontranslations
    \draw (0,2pt) -- (0,0) node [below] {\scriptsize $\x$};
  \end{scope}
}
\foreach \y in {0.0, 1.0, 2.0}
{
  \begin{scope}[xshift=-3cm,yshift=\y cm]
    \pgftransformresetnontranslations
    \draw (2pt,0) -- (0,0) node [left] {\scriptsize $\y$};
  \end{scope}
}

\begin{scope}
\clip (-3,0) rectangle (3, 2);
% beta = -inf
\draw [PlotColorA,thick] plot [smooth] coordinates {(-3,0.989) (-2.667,0.971)
(-2.5,0.956) (-2.333,0.934) (-2.083,0.886) (-1.917,0.841) (-1.833,0.814)
(-1.667,0.751) (-1.417,0.633) (-1.167,0.494) (-0.75,0.245) (-0.583,0.156)
(-0.5,0.118) (-0.333,0.054) (-0.25,0.031) (-0.167,0.014) (-0.083,0.003) (0,0)
(0.083,0.003) (0.167,0.014) (0.25,0.031) (0.333,0.054) (0.5,0.118) (0.583,0.156)
(0.75,0.245) (1.167,0.494) (1.417,0.633) (1.667,0.751) (1.833,0.814)
(1.917,0.841) (2.083,0.886) (2.333,0.934) (2.5,0.956) (2.667,0.971) (3,0.989)};
% beta = -2
\draw [PlotColorB,thick] plot [smooth] coordinates {(-3,1.385) (-2.75,1.308)
(-2.5,1.22) (-2.25,1.117) (-2,1) (-1.833,0.913) (-1.583,0.771) (-1.333,0.615)
(-0.917,0.347) (-0.75,0.247) (-0.583,0.157) (-0.5,0.118) (-0.333,0.054)
(-0.25,0.031) (-0.167,0.014) (-0.083,0.003) (0,0) (0.083,0.003) (0.167,0.014)
(0.25,0.031) (0.333,0.054) (0.5,0.118) (0.583,0.157) (0.75,0.247) (0.917,0.347)
(1.333,0.615) (1.583,0.771) (1.833,0.913) (2,1) (2.25,1.117) (2.5,1.22)
(2.75,1.308) (3,1.385)};
% beta = -1
\draw [PlotColorC,thick] plot [smooth] coordinates {(-3,1.5) (-2.75,1.401)
(-2.5,1.292) (-2.25,1.17) (-2,1.036) (-1.833,0.94) (-1.583,0.786) (-0.833,0.297)
(-0.583,0.157) (-0.417,0.083) (-0.333,0.054) (-0.25,0.031) (-0.167,0.014)
(-0.083,0.003) (0,0) (0.083,0.003) (0.167,0.014) (0.25,0.031) (0.333,0.054)
(0.417,0.083) (0.583,0.157) (0.833,0.297) (1.583,0.786) (1.833,0.94) (2,1.036)
(2.25,1.17) (2.5,1.292) (2.75,1.401) (3,1.5)};
% beta = 0
\draw [PlotColorD,thick] plot [smooth] coordinates {(-3,1.705) (-2.583,1.467)
(-2.167,1.208) (-1.75,0.929) (-1.083,0.462) (-0.917,0.351) (-0.75,0.248)
(-0.583,0.157) (-0.5,0.118) (-0.333,0.054) (-0.25,0.031) (-0.167,0.014)
(-0.083,0.003) (0,0) (0.083,0.003) (0.167,0.014) (0.25,0.031) (0.333,0.054)
(0.5,0.118) (0.583,0.157) (0.75,0.248) (0.917,0.351) (1.083,0.462) (1.75,0.929)
(2.167,1.208) (2.583,1.467) (3,1.705)};
% beta = 0.5
\draw [PlotColorE,thick] plot [smooth] coordinates {(-3,1.88) (-2.333,1.401)
(-1.25,0.586) (-1,0.409) (-0.833,0.299) (-0.667,0.201) (-0.583,0.157)
(-0.417,0.083) (-0.333,0.054) (-0.25,0.031) (-0.167,0.014) (-0.083,0.003) (0,0)
(0.083,0.003) (0.167,0.014) (0.25,0.031) (0.333,0.054) (0.417,0.083)
(0.583,0.157) (0.667,0.201) (0.833,0.299) (1,0.409) (1.25,0.586) (2.333,1.401)
(3,1.88)};
% beta = 1
\draw [PlotColorF,thick] plot [smooth] coordinates {(-3,2.162) (-2.333,1.539)
(-1.833,1.088) (-1.417,0.734) (-1.167,0.537) (-1,0.414) (-0.833,0.302)
(-0.667,0.202) (-0.5,0.118) (-0.417,0.083) (-0.333,0.054) (-0.25,0.031)
(-0.167,0.014) (-0.083,0.003) (0,0) (0.083,0.003) (0.167,0.014) (0.25,0.031)
(0.333,0.054) (0.417,0.083) (0.5,0.118) (0.667,0.202) (0.833,0.302) (1,0.414)
(1.167,0.537) (1.417,0.734) (1.833,1.088) (2.333,1.539) (3,2.162)};
% beta = 2
\draw [PlotColorG,thick] plot [smooth] coordinates {(-3,4.5) (-2.833,4.014)
(-2.583,3.337) (-2.417,2.92) (-2.25,2.531) (-2.083,2.17) (-1.833,1.681)
(-1.667,1.389) (-1.5,1.125) (-1.417,1.003) (-1.25,0.781) (-1.083,0.587)
(-0.917,0.42) (-0.75,0.281) (-0.583,0.17) (-0.5,0.125) (-0.417,0.087)
(-0.333,0.056) (-0.25,0.031) (-0.167,0.014) (-0.083,0.003) (0,0) (0.083,0.003)
(0.167,0.014) (0.25,0.031) (0.333,0.056) (0.417,0.087) (0.5,0.125) (0.583,0.17)
(0.75,0.281) (0.917,0.42) (1,0.5) (1.167,0.681) (1.333,0.889) (1.5,1.125)
(1.667,1.389) (1.833,1.681) (2.083,2.17) (2.25,2.531) (2.417,2.92) (2.583,3.337)
(2.75,3.781) (3,4.5)};
\end{scope}

\end{tikzpicture}

%% file: figures/compare_huber.tikz
% TikZ Figure.

\begin{tikzpicture}[scale=1]

\draw (0,1.0) node [above] {\small Original\vphantom{y}};
\draw (0,0) node {%
\includegraphics[width=3.2cm]{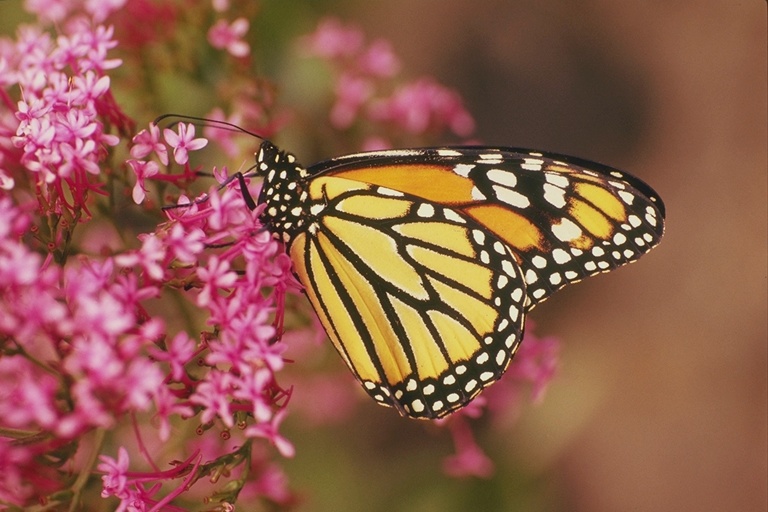}};
\draw (0,-2.3) node {%
\includegraphics[width=2.2cm]{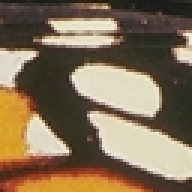}};

\begin{scope}[xshift=3.3cm]
\draw (0,1.0) node [above] {\small Noisy\vphantom{y}};
\draw (0,0) node {%
\includegraphics[width=3.2cm]{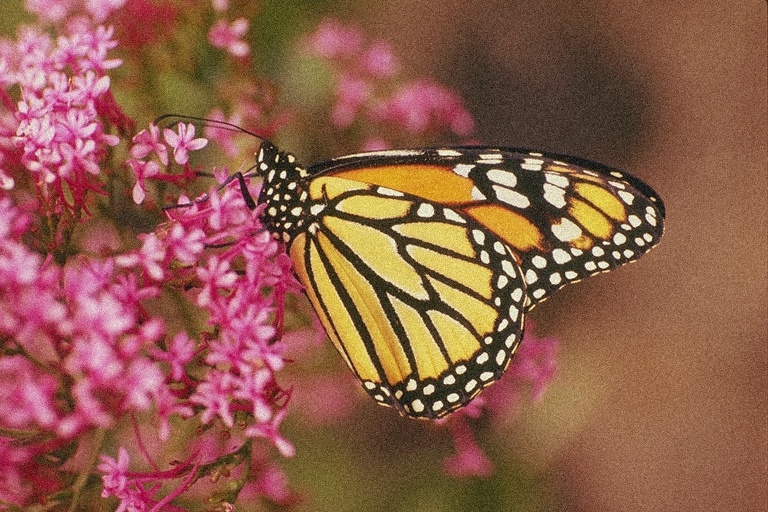}};
\draw (0,-2.3) node {%
\includegraphics[width=2.2cm]{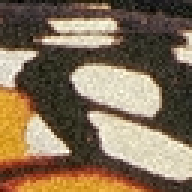}};
\end{scope}

\begin{scope}[xshift=6.6cm]
\draw (0,1.0) node [above] {\small TV\vphantom{y}};
\draw (0,0) node {%
\includegraphics[width=3.2cm]{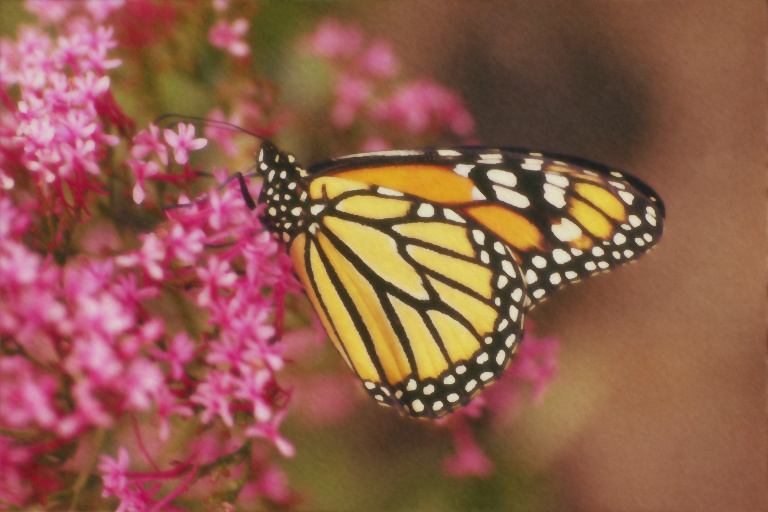}};
\draw (0,-2.3) node {%
\includegraphics[width=2.2cm]{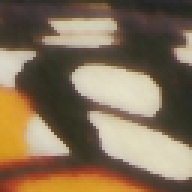}};
\end{scope}

\begin{scope}[xshift=9.9cm]
\draw (0,1.0) node [above] {\small First-order Approx.\vphantom{y}};
\draw (0,0) node {%
\includegraphics[width=3.2cm]{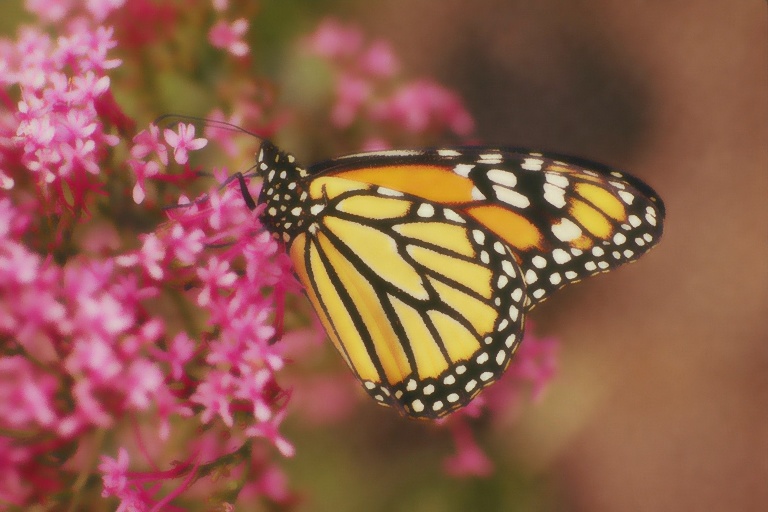}};
\draw (0,-2.3) node {%
\includegraphics[width=2.2cm]{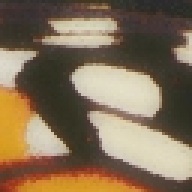}};
\end{scope}

\begin{scope}[xshift=13.2cm]
\draw (0,1.0) node [above] {\small Second-order Approx.\vphantom{y}};
\draw (0,0) node {%
\includegraphics[width=3.2cm]{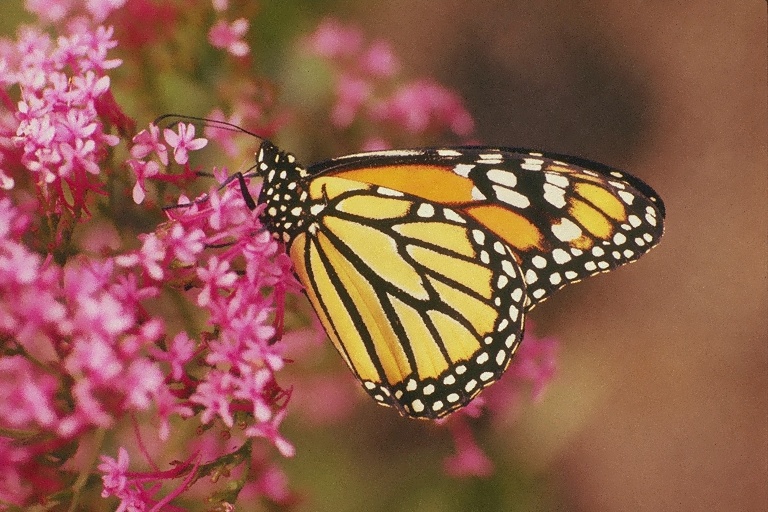}};
\draw (0,-2.3) node {%
\includegraphics[width=2.2cm]{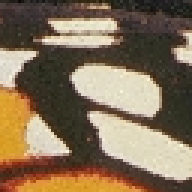}};
\end{scope}

\end{tikzpicture}

%% file: figures/compare_gaussian.tikz
% TikZ Figure.

\begin{tikzpicture}[scale=1]

\draw (0,1.3) node [above] {\small Original\vphantom{y}};
\draw (0,0) node {%
\includegraphics[width=4cm]{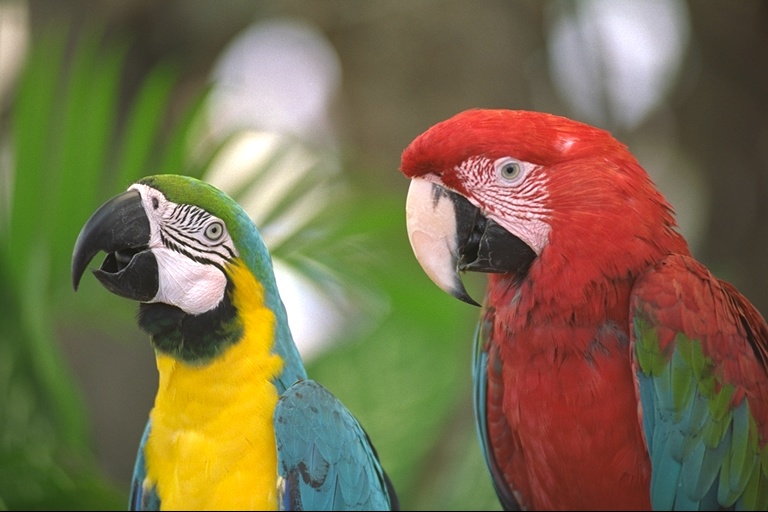}};
\draw (0,-2.90) node {%
\includegraphics[width=3cm]{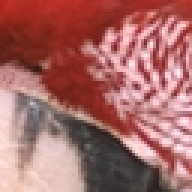}};

\begin{scope}[xshift=4.1cm]
\draw (0,1.3) node [above] {\small Bilateral Filtered\vphantom{y}};
\draw (0,0) node {%
\includegraphics[width=4cm]{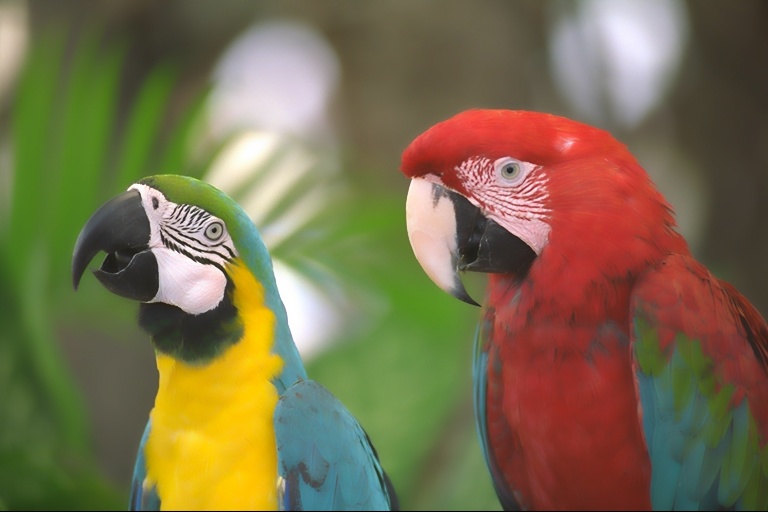}};
\draw (0,-2.90) node {%
\includegraphics[width=3cm]{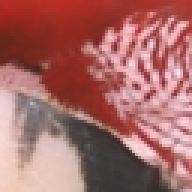}};
\end{scope}

\begin{scope}[xshift=8.2cm]
\draw (0,1.3) node [above] {\small First-order Approx.\vphantom{y}};
\draw (0,0) node {%
\includegraphics[width=4cm]{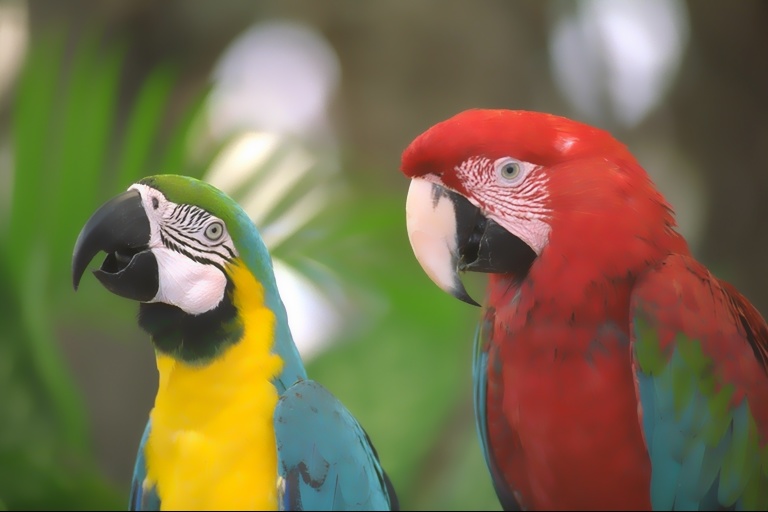}};
\draw (0,-2.90) node {%
\includegraphics[width=3cm]{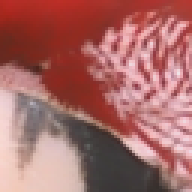}};
\end{scope}

\begin{scope}[xshift=12.3cm]
\draw (0,1.3) node [above] {\small Second-order Approx.\vphantom{y}};
\draw (0,0) node {%
\includegraphics[width=4cm]{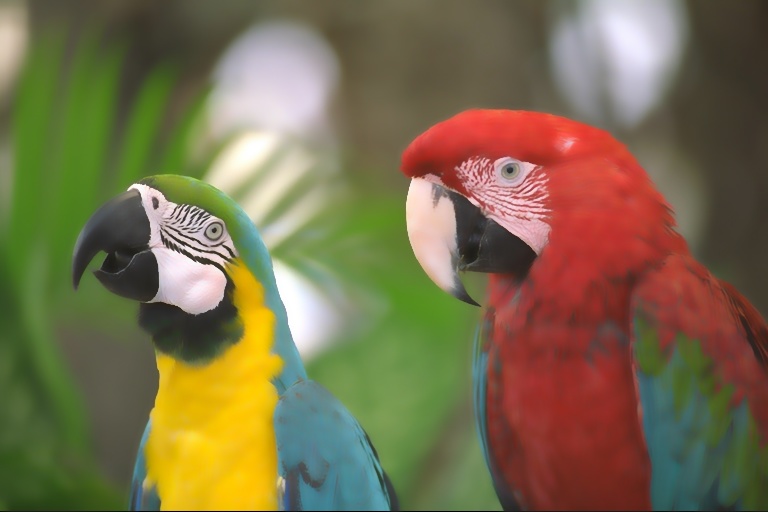}};
\draw (0,-2.90) node {%
\includegraphics[width=3cm]{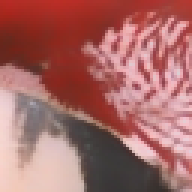}};
\end{scope}

\end{tikzpicture}

%% file: figures/compare_boxcar.tikz
% TikZ Figure.

\begin{tikzpicture}[scale=1]

\draw (0,1.3) node [above] {\small Original\vphantom{y}};
\draw (0,0) node {%
\includegraphics[width=4cm]{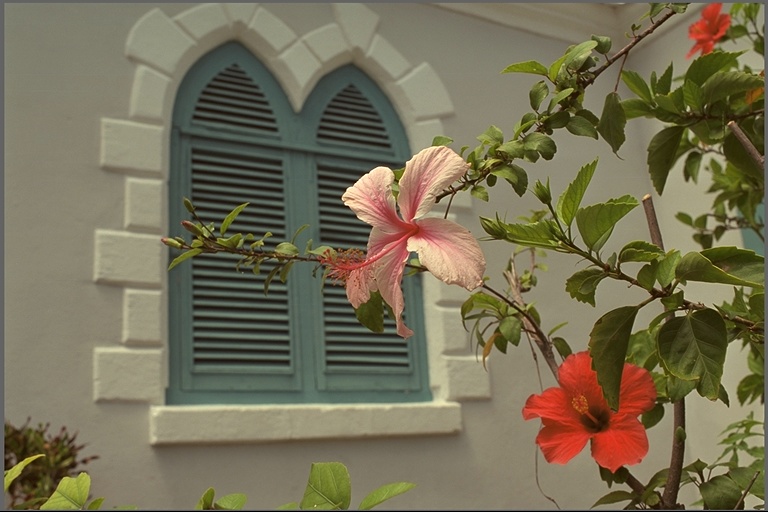}};
\draw (0,-2.90) node {%
\includegraphics[width=3cm]{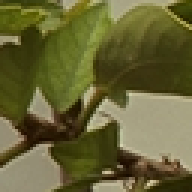}};

\begin{scope}[xshift=4.1cm]
\draw (0,1.3) node [above] {\small Bilateral Filtered\vphantom{y}};
\draw (0,0) node {%
\includegraphics[width=4cm]{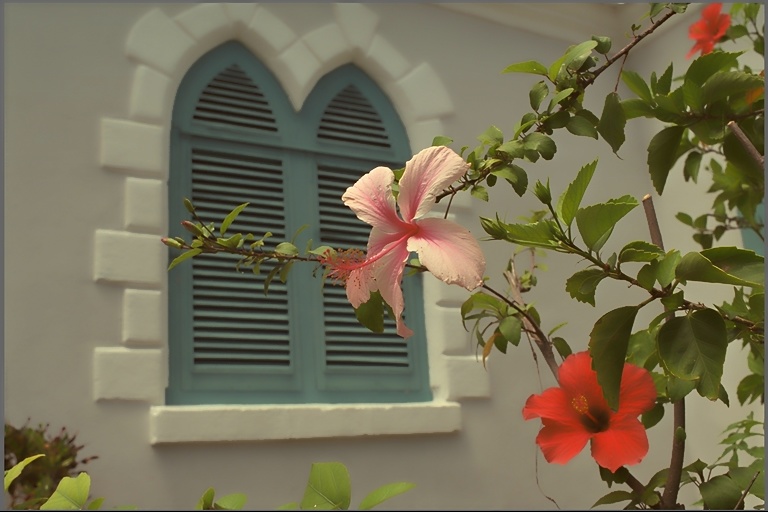}};
\draw (0,-2.90) node {%
\includegraphics[width=3cm]{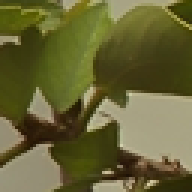}};
\end{scope}

\begin{scope}[xshift=8.2cm]
\draw (0,1.3) node [above] {\small First-order Approx.\vphantom{y}};
\draw (0,0) node {%
\includegraphics[width=4cm]{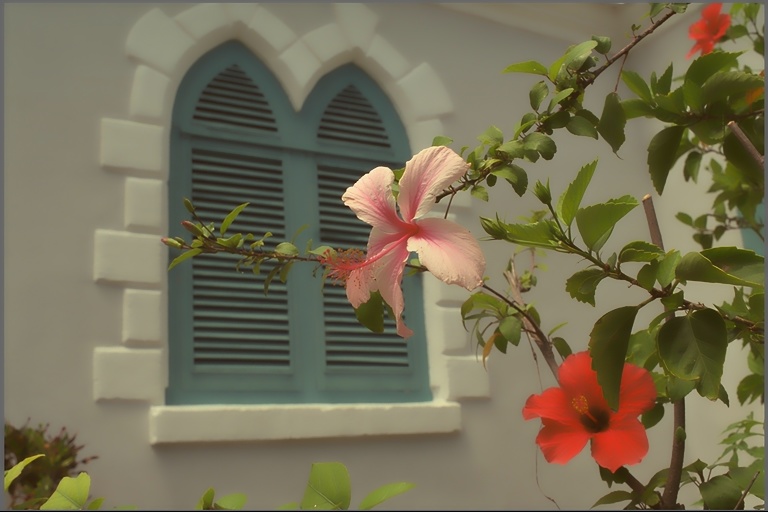}};
\draw (0,-2.90) node {%
\includegraphics[width=3cm]{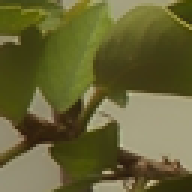}};
\end{scope}

\begin{scope}[xshift=12.3cm]
\draw (0,1.3) node [above] {\small Second-order Approx.\vphantom{y}};
\draw (0,0) node {%
\includegraphics[width=4cm]{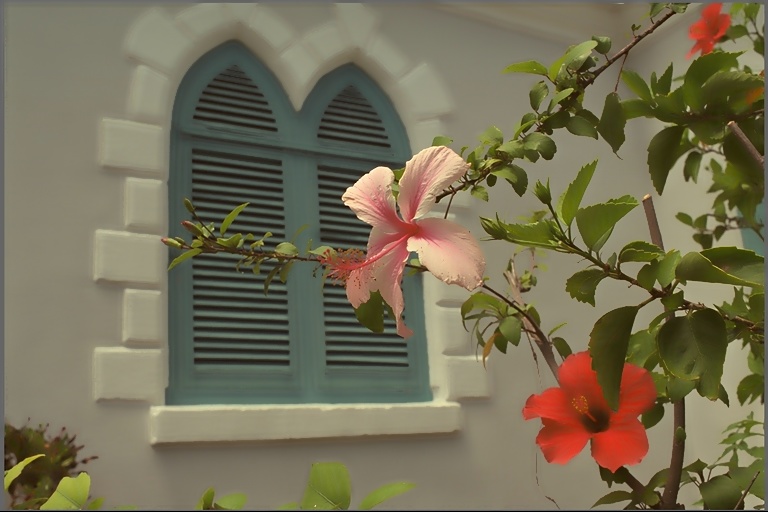}};
\draw (0,-2.90) node {%
\includegraphics[width=3cm]{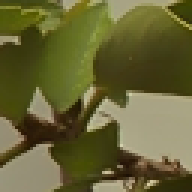}};
\end{scope}

\end{tikzpicture}

%% file: figures/compare_exponential.tikz
% TikZ Figure.

\begin{tikzpicture}[scale=1]

\draw (0,2.9) node [above] {\small Original\vphantom{y}};
\draw (0,0) node {%
\includegraphics[width=4cm]{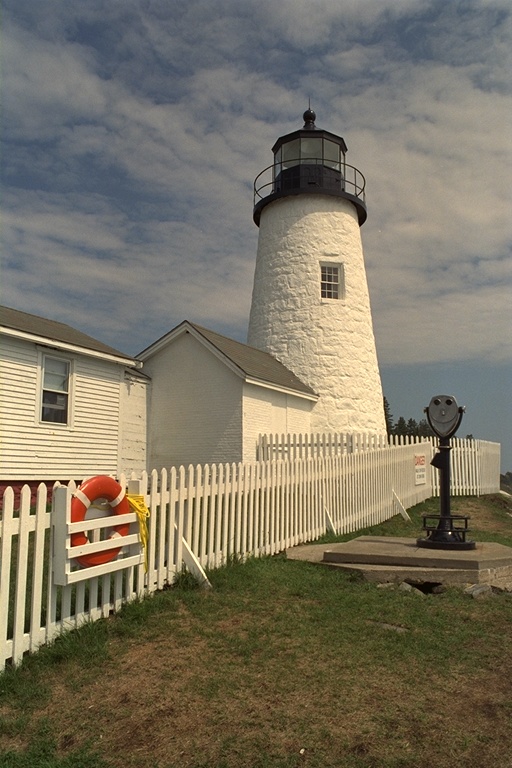}};
\draw (0,-4.6) node {%
\includegraphics[width=3cm]{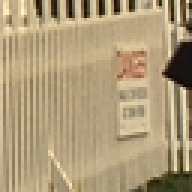}};

\begin{scope}[xshift=4.1cm]
\draw (0,2.9) node [above] {\small Bilateral Filtered\vphantom{y}};
\draw (0,0) node {%
\includegraphics[width=4cm]{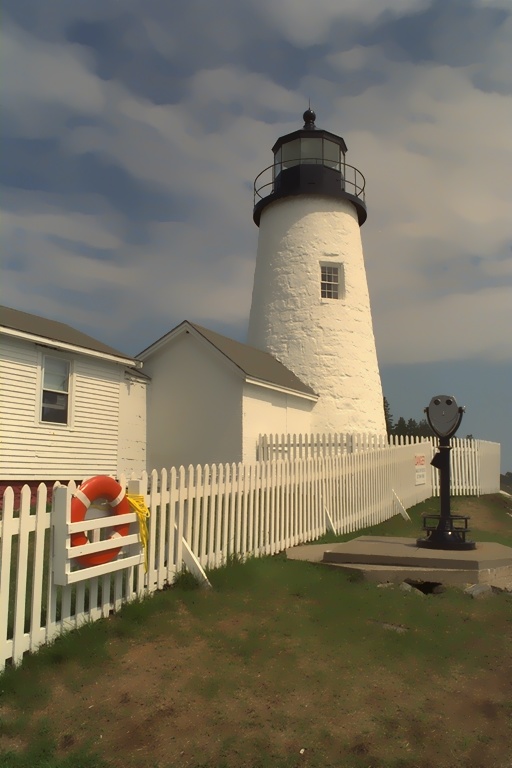}};
\draw (0,-4.6) node {%
\includegraphics[width=3cm]{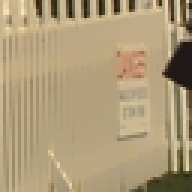}};
\end{scope}

\begin{scope}[xshift=8.2cm]
\draw (0,2.9) node [above] {\small First-order Approx.\vphantom{y}};
\draw (0,0) node {%
\includegraphics[width=4cm]{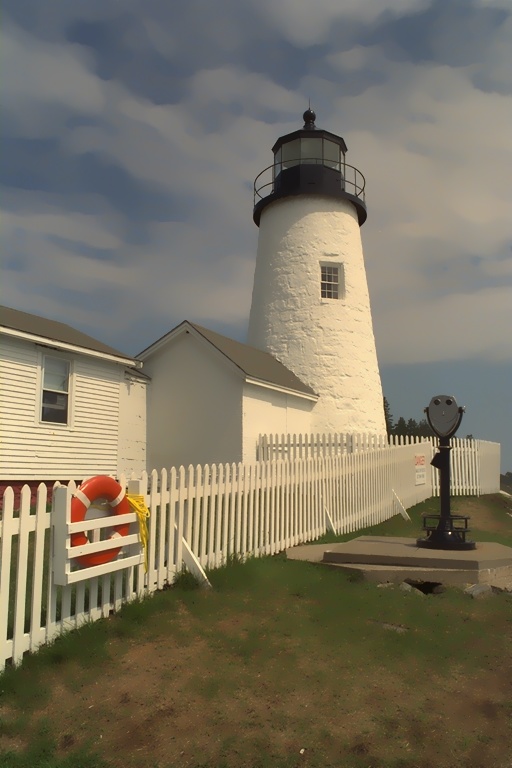}};
\draw (0,-4.6) node {%
\includegraphics[width=3cm]{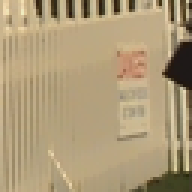}};
\end{scope}

\begin{scope}[xshift=12.3cm]
\draw (0,2.9) node [above] {\small Second-order Approx.\vphantom{y}};
\draw (0,0) node {%
\includegraphics[width=4cm]{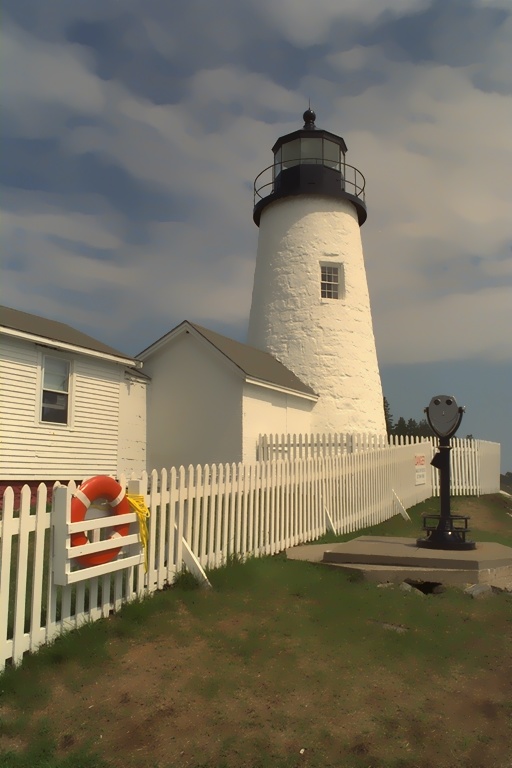}};
\draw (0,-4.6) node {%
\includegraphics[width=3cm]{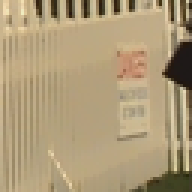}};
\end{scope}

\end{tikzpicture}

%% file: figures/h1_distance.tikz
% TikZ Figure: Distance between exact and approx filters for H1.

\begin{tikzpicture}[scale=1.35,xscale=7]
\draw (0,2.3) -- (0,0)
  node [midway,rotate=90,yshift=30pt]
  {\footnotesize $\|\vv{w} - \vv{w}_\mathit{approx}\|_1$}
  -- (0.6,0) node [midway,below=11pt]
  {\footnotesize $\sigma$};
\foreach \x in {0.0, 0.2, 0.4, 0.6}
{
  \begin{scope}[xshift=\x cm]
    \pgftransformresetnontranslations
    \draw (0,2pt) -- (0,0) node [below] {\scriptsize $\x$};
  \end{scope}
}
\foreach \y in {0.0, 1.0, 2.0}
{
  \begin{scope}[xshift=0cm,yshift=\y cm]
    \pgftransformresetnontranslations
    \draw (2pt,0) -- (0,0) node [left] {\scriptsize $\y$};
  \end{scope}
}
\draw [PlotColorA] plot [smooth] coordinates {(0,0) (0.0122,0) (0.0245,0)
  (0.0367,0.0001) (0.0490,0.0004) (0.0612,0.0009) (0.0735,0.0018)
  (0.0857,0.0033) (0.0980,0.0055) (0.1102,0.0087) (0.1224,0.0129)
  (0.1347,0.0185) (0.1469,0.0257) (0.1592,0.0345) (0.1714,0.0453)
  (0.1837,0.0581) (0.1959,0.0732) (0.2082,0.0907) (0.2204,0.1106)
  (0.2327,0.1333) (0.2449,0.1587) (0.2571,0.1869) (0.2694,0.2180)
  (0.2816,0.2521) (0.2939,0.2893) (0.3061,0.3295) (0.3184,0.3728)
  (0.3306,0.4193) (0.3429,0.4688) (0.3551,0.5216) (0.3673,0.5774)
  (0.3796,0.6365) (0.3918,0.6986) (0.4041,0.7639) (0.4163,0.8323)
  (0.4286,0.9037) (0.4408,0.9783) (0.4531,1.0559) (0.4653,1.1365)
  (0.4776,1.2201) (0.4898,1.3068) (0.5020,1.3963) (0.5143,1.4889)
  (0.5265,1.5843) (0.5388,1.6826) (0.5510,1.7838) (0.5633,1.8879)
  (0.5755,1.9947) (0.5878,2.1044) (0.6000,2.2169)};
\end{tikzpicture}

%% file: paper.bbl
\begin{thebibliography}{10}
\providecommand{\url}[1]{#1}
\csname url@samestyle\endcsname
\providecommand{\newblock}{\relax}
\providecommand{\bibinfo}[2]{#2}
\providecommand{\BIBentrySTDinterwordspacing}{\spaceskip=0pt\relax}
\providecommand{\BIBentryALTinterwordstretchfactor}{4}
\providecommand{\BIBentryALTinterwordspacing}{\spaceskip=\fontdimen2\font plus
\BIBentryALTinterwordstretchfactor\fontdimen3\font minus
  \fontdimen4\font\relax}
\providecommand{\BIBforeignlanguage}[2]{{%
\expandafter\ifx\csname l@#1\endcsname\relax
\typeout{** WARNING: IEEEtran.bst: No hyphenation pattern has been}%
\typeout{** loaded for the language `#1'. Using the pattern for}%
\typeout{** the default language instead.}%
\else
\language=\csname l@#1\endcsname
\fi
#2}}
\providecommand{\BIBdecl}{\relax}
\BIBdecl

\bibitem{smith1997susan}
S.~M. Smith and J.~M. Brady, ``{SUSAN-A new approach to low level image
  processing},'' \emph{International journal of computer vision}, vol.~23,
  no.~1, pp. 45--78, 1997.

\bibitem{tomasi1998bilateral}
C.~Tomasi and R.~Manduchi, ``Bilateral filtering for gray and color images,''
  in \emph{Computer Vision, 1998. Sixth International Conference on}.\hskip 1em
  plus 0.5em minus 0.4em\relax IEEE, 1998, pp. 839--846.

\bibitem{buades2005review}
A.~Buades, B.~Coll, and J.-M. Morel, ``A review of image denoising algorithms,
  with a new one,'' \emph{Multiscale Modeling \& Simulation}, vol.~4, no.~2,
  pp. 490--530, 2005.

\bibitem{milanfar2013tour}
P.~Milanfar, ``A tour of modern image filtering: New insights and methods, both
  practical and theoretical,'' \emph{IEEE Signal Processing Magazine}, vol.~30,
  no.~1, pp. 106--128, 2013.

\bibitem{talebi2014nonlocal}
H.~Talebi and P.~Milanfar, ``Nonlocal image editing,'' \emph{IEEE Transactions
  on Image Processing}, vol.~23, no.~10, pp. 4460--4473, 2014.

\bibitem{protter2009generalizing}
M.~Protter, M.~Elad, H.~Takeda, and P.~Milanfar, ``Generalizing the
  nonlocal-means to super-resolution reconstruction,'' \emph{IEEE Transactions
  on image processing}, vol.~18, no.~1, pp. 36--51, 2009.

\bibitem{buades09}
A.~Buades, B.~Coll, J.~M. Morel, and C.~Sbert, ``Self-similarity driven color
  demosaicking,'' \emph{IEEE Transactions on Image Processing}, vol.~18, no.~6,
  pp. 1192--1202, June 2009.

\bibitem{isidoro2016pull}
J.~Isidoro and P.~Milanfar, ``A pull-push method for fast non-local means
  filtering,'' in \emph{IEEE International Conference on Image Processing},
  2016.

\bibitem{talebi2016fast}
H.~Talebi and P.~Milanfar, ``Fast multilayer laplacian enhancement,''
  \emph{IEEE Transactions on Computational Imaging}, vol.~2, no.~4, pp.
  496--509, 2016.

\bibitem{elad2002origin}
M.~Elad, ``On the origin of the bilateral filter and ways to improve it,''
  \emph{IEEE Transactions on image processing}, vol.~11, no.~10, pp.
  1141--1151, 2002.

\bibitem{takeda07}
H.~Takeda, S.~Farsiu, and P.~Milanfar, ``Kernel regression for image processing
  and reconstruction,'' \emph{IEEE Transactions on Image Processing}, vol.~16,
  no.~2, pp. 349--366, Feb 2007.

\bibitem{durand2002fast}
F.~Durand and J.~Dorsey, ``Fast bilateral filtering for the display of
  high-dynamic-range images,'' in \emph{ACM transactions on graphics (TOG)},
  vol.~21, no.~3.\hskip 1em plus 0.5em minus 0.4em\relax ACM, 2002, pp.
  257--266.

\bibitem{black1998robust}
M.~J. Black, G.~Sapiro, D.~H. Marimont, and D.~Heeger, ``Robust anisotropic
  diffusion,'' \emph{IEEE Transactions on Image Processing}, vol.~7, no.~3, pp.
  421--432, Mar 1998.

\bibitem{louchet2011total}
C.~Louchet and L.~Moisan, ``Total variation as a local filter,'' \emph{SIAM
  Journal on Imaging Sciences}, vol.~4, no.~2, pp. 651--694, 2011.

\bibitem{plugandplay}
S.~V. Venkatakrishnan, C.~A. Bouman, and B.~Wohlberg, ``Plug-and-play priors
  for model based reconstruction,'' in \emph{2013 IEEE Global Conference on
  Signal and Information Processing}, Dec 2013, pp. 945--948.

\bibitem{meinhardt17learning}
T.~Meinhardt, M.~Möller, C.~Hazirbas, and D.~Cremers, ``Learning proximal
  operators: Using denoising networks for regularizing inverse imaging
  problems,'' in \emph{ICCV}, October 2017.

\bibitem{Combettes2011}
P.~L. Combettes and J.-C. Pesquet, \emph{Proximal Splitting Methods in Signal
  Processing}.\hskip 1em plus 0.5em minus 0.4em\relax New York, NY: Springer
  New York, 2011, pp. 185--212.

\bibitem{condat2014}
L.~Condat, ``A generic proximal algorithm for convex optimization: Application
  to total variation minimization,'' \emph{IEEE Signal Processing Letters},
  vol.~21, no.~8, pp. 985--989, Aug 2014.

\bibitem{romano2016little}
Y.~Romano, M.~Elad, and P.~Milanfar, ``{The little engine that could:
  Regularization by denoising (RED)},'' \emph{SIAM Journal on Imaging Science},
  vol.~10, no.~4, pp. 1804--1844, 2017.

\bibitem{chatterjee11}
P.~Chatterjee and P.~Milanfar, ``Practical bounds on image denoising: From
  estimation to information,'' \emph{IEEE Trans. on Image Processing}, vol.~20,
  no.~5, pp. 1221--1233, 2011.

\bibitem{talebi2013saif}
H.~Talebi, X.~Zhu, and P.~Milanfar, ``How to saif-ly boost denoising
  performance,'' \emph{IEEE Transactions on Image Processing}, vol.~22, no.~4,
  pp. 1470--1485, 2013.

\bibitem{milanfar2013symmetrizing}
P.~Milanfar, ``Symmetrizing smoothing filters,'' \emph{SIAM, Journal on Imaging
  Science}, vol.~6, no.~1, pp. 263--284, 2013.

\bibitem{chatterjee2010denoising}
P.~Chatterjee and P.~Milanfar, ``Is denoising dead?'' \emph{IEEE Transactions
  on Image Processing}, vol.~19, no.~4, pp. 895--911, 2010.

\bibitem{proximal2015}
N.~Polson, J.~G. Scott, and B.~T. Willard, ``Proximal algorithms in statistics
  and machine learning,'' \emph{Statistical Science}, vol.~30, no.~4, pp.
  559--581, 2015.

\bibitem{robbins56}
H.~Robbins, ``An empirical bayes approach to statistics,'' \emph{Third Berkeley
  Statistics Symposium}, 1956.

\bibitem{stein81}
C.~Stein, ``Estimation of the mean of a multivariate normal distribution,''
  \emph{Annals of Statistics}, vol.~9, no.~6, 1981.

\bibitem{efron11}
B.~Efron, ``Tweedie’s formula and selection bias,'' \emph{Journal of the
  American Statistical Association}, vol. 106, no. 496, pp. 1602--1614, 2011.

\bibitem{elmoataz2008}
\BIBentryALTinterwordspacing
A.~Elmoataz, O.~Lezoray, S.~Bougleux, and V.~T. Ta, ``{Unifying local and
  nonlocal processing with partial difference operators on weighted graphs},''
  in \emph{{International Workshop on Local and Non-Local Approximation in
  Image Processing}}, Switzerland, 2008, pp. 11--26. [Online]. Available:
  \url{https://hal.archives-ouvertes.fr/hal-00329521}
\BIBentrySTDinterwordspacing

\bibitem{huber1981}
P.~J. Huber, \emph{Robust Statistics}.\hskip 1em plus 0.5em minus 0.4em\relax
  Wiley, 1981.

\bibitem{james1961}
\BIBentryALTinterwordspacing
W.~James and C.~Stein, ``Estimation with quadratic loss,'' in \emph{Proceedings
  of the Fourth Berkeley Symposium on Mathematical Statistics and Probability,
  Volume 1: Contributions to the Theory of Statistics}.\hskip 1em plus 0.5em
  minus 0.4em\relax Berkeley, Calif.: University of California Press, 1961, pp.
  361--379. [Online]. Available:
  \url{https://projecteuclid.org/euclid.bsmsp/1200512173}
\BIBentrySTDinterwordspacing

\bibitem{park91}
J.~Park and I.~Sandberg, ``Universal approximation using radial-basis-function
  networks,'' \emph{Neural Computation}, vol.~3, no.~2, pp. 246--257, 1991.

\bibitem{vert04}
J.~Vert, K.~Tsuda, and B.~Sch{\"o}lkopf, \emph{A Primer on Kernel
  Methods}.\hskip 1em plus 0.5em minus 0.4em\relax Cambridge, MA, USA: MIT
  Press, 2004, pp. 35--70.

\bibitem{magnus99}
J.~R. Magnus and H.~Neudecker, \emph{Matrix Differential Calculus with
  Applications in Statistics and Econometrics}, 2nd~ed.\hskip 1em plus 0.5em
  minus 0.4em\relax John Wiley, 1999.

\bibitem{rahimi07}
A.~Rahimi and B.~Recht, ``Random features for large-scale kernel machines,''
  \emph{Advances in Neural Information Processing Systems}, 2007.

\bibitem{genton01}
M.~Genton, ``Classes of kernels for machine learning: A statistical
  perspective,'' \emph{Journal of Machine Learning Research}, vol.~2, pp.
  299--312, 2001.

\bibitem{milanfar2016new}
P.~Milanfar and H.~Talebi, ``A new class of image filters without
  normalization,'' in \emph{Image Processing (ICIP), 2016 IEEE International
  Conference on}.\hskip 1em plus 0.5em minus 0.4em\relax IEEE, 2016, pp.
  3294--3298.

\bibitem{welsch1977}
P.~W. Holland and R.~E. Welsch, ``Robust regression using iteratively
  reweighted least-squares,'' \emph{Communications in Statistics - Theory and
  Methods}, vol.~6, no.~9, pp. 813--827, 1977.

\bibitem{barron2017more}
J.~T. Barron, ``A more general robust loss function,'' \emph{arXiv preprint
  arXiv:1701.03077}, 2017.

\bibitem{spivak1970comprehensive}
M.~Spivak, \emph{A comprehensive introduction to differential geometry}, ser. A
  Comprehensive Introduction to Differential Geometry.\hskip 1em plus 0.5em
  minus 0.4em\relax Brandeis University, 1970, no. v. 1.

\bibitem{alexandrov39}
A.~Alexandrov, ``Almost everywhere existence of the second differential of a
  convex function and some properties of convex surfaces connected with it,''
  \emph{Leningrad State Univ. Annals [Uchenye Zapiski]}, 1939.

\bibitem{gutierrez2016monge}
C.~E. Guti{\'e}rrez, \emph{The Monge--Ampere equation}.\hskip 1em plus 0.5em
  minus 0.4em\relax Springer, 2016.

\bibitem{polyak1964some}
B.~T. Polyak, ``Some methods of speeding up the convergence of iteration
  methods,'' \emph{USSR Computational Mathematics and Mathematical Physics},
  vol.~4, no.~5, pp. 1--17, 1964.

\end{thebibliography}
